%% file: main.tex
\documentclass[letterpaper, 10 pt, conference]{ieeeconf}  

\IEEEoverridecommandlockouts                              

\usepackage{graphicx} 
\usepackage{mathtools}
\usepackage{epsfig} 
\usepackage{amssymb}  
\usepackage{subfig}
\usepackage{pgfplots}
\usepackage{tikz}

\newlength\figureheight
\newlength\figurewidth

\pgfplotsset{compat=newest}

\title{\LARGE \bf
Online Human Gesture Recognition using Recurrent Neural Networks and Wearable Sensors}

\author{Alessandro Carf\`i \and Carola Motolese \and Barbara Bruno \and Fulvio Mastrogiovanni 
\thanks{A. Carf\`i, C. Motolese, B. Bruno and F. Mastrogiovanni are with the Department of Informatics, Bioengineering, Robotics, and Systems Engineering, University of Genoa, Via Opera Pia 13, 16145 Genoa, Italy.
	Corresponding author's email:
	alessandro.carfi@dibris.unige.it.}
}

\begin{document}

\twocolumn[{
\copyright 2018 IEEE Personal use of this material is permitted. Permission from IEEE must be obtained for all other uses, in any current or future media, including reprinting/republishing this material for advertising or promotional purposes, creating new collective works, for resale or redistribution to servers or lists, or reuse of any copyrighted component of this work in other works.\\
	
This work has been published in the proceedings of the 2018 IEEE International Conference on Robot and Human Interactive Communication (RO-MAN), 27-31 August, Nanjing, China.}]

\maketitle
\thispagestyle{empty}
\pagestyle{empty}

\begin{abstract}
	Gestures are a natural communication modality for humans. The ability to interpret gestures is fundamental for robots aiming to naturally interact with humans. Wearable sensors are promising to monitor human activity, in particular the usage of triaxial accelerometers for gesture recognition have been explored. Despite this, the state of the art presents lack of systems for reliable online gesture recognition using accelerometer data. The article proposes SLOTH, an architecture for online gesture recognition, based on a wearable triaxial accelerometer, a Recurrent Neural Network (RNN) probabilistic classifier and a procedure for continuous gesture detection, relying on modelling gesture probabilities, that guarantees (i) good recognition results in terms of precision and recall, (ii) immediate system reactivity. 
\end{abstract}
\section{Introduction}
\label{sec:introduction}
\noindent
Gestures are an intuitive and natural communication modality that humans use daily to convey information, intentionally (communicative gestures) or un-intentionally (informative gestures) \cite{lyons1977semantics}. Flight attendants indicating emergency exits during the pre-flight safety demonstration, the aircraft marshaller that uses hand and body gestures to direct flight operations, deaf-mutes using sign language to communicate and infants referring to unknown objects by pointing them, are all examples of communicative gestures usage. Instead, a lady lifting a glass to her mouth to drink, informs those who are observing her that she is thirsty (informative gesture).

The ability to recognise gestures and discriminate between communicative and informative ones is of great importance for robots interacting or collaborating with humans \cite{darvish2018flexible}. Two possible application scenarios are the smart home and the smart factory one. In a smart home humans could use communicative gestures to control appliances or interact with a robot companion, while informative gestures could be used to recognise occurrences of specific daily living activities or, especially when dealing with elderly users, to monitor their health status. Similarly, gesture based protocol can be used in smart-factories to enhance the interaction between human operators and robot co-workers. 
Although the difference between communicative and informative gesture is fundamental to understand the \textit{meaning} of a gesture, in our current formulation of the gesture recognition problem we decided not to address it, and leave it for future work.

The gesture recognition problem can be divided into three sub-problems: acquire informative data (\textit{perception}), establish which portion of data refers to a gesture (\textit{detection}) and determine the class the detected portion of data belongs (\textit{classification}). A common approach to the human gesture perception involves the usage of vision-based systems relaying either on RGB \cite{yang2007gesture} or RGB-D \cite{iengo2014continuous} cameras. Vision-based techniques have many drawbacks such as: need of structured environment, high computation complexity and sensitivity to partial occlusion. Instead wearable triaxial accelerometers provide sufficient information to perceive human movements in a non-invasive way, since they can be incorporated in everyday objects such as watches, wristbands or clothing.

The main contribution of the article is a recognition procedure, that we refer to as SLOTH, for hand/arm gesture perceived by smartwatch accelerometer. SLOTH relies on a Recurrent Neural Network (RNN) probabilistic classifier and a novel algorithm that processes the instantaneous probabilities associated with each gesture classes, generated by the RNN module, to continuously detect and classify gestures occurrences.

%

The paper is organised as follows. Section \ref{sec:background} gives a brief overview of gesture recognition approaches relying on inertial information. Section \ref{sec:sofar} describes the proposed method for continuous gesture recognition, while details about the implementation are discussed in Section \ref{sec:implementation}. Section \ref{sec:results} discusses the experimental evaluation. Conclusions follow. 
\section{Background}
\label{sec:background}
\noindent
Literature shows that different approaches have been proposed to tackle the problem of gesture detection and classification using wearable inertial sensors. Usually, the two problems are studied separately to reduce their complexity. 

While there is no standard work-flow to solve the classification problem, most solutions implement four key steps: \textit{data preprocessing}, \textit{feature extraction}, \textit{model building} and \textit{classification}. Accelerometer data are typically affected by high-frequency noise that can be filtered out using different techniques such as moving average filters \cite{xie2016accelerometer} \cite{khan2013exploratory} \cite{moazen2016airdraw}, median filters \cite{bruno2014using}, temporal compression \cite{akl2010accelerometer}, quantisation \cite{liu2009uwave} or Hanning filters \cite{wu2010hand}. Accelerometers measure the proper acceleration of the object they are attached to, which includes the gravity acceleration and any other acceleration that the object is subject to (in the case of wearable sensors, any other acceleration produced by a person's movements). The gravity acceleration can be used as an independent source of information for the classification \cite{bruno2014using}, to isolate body acceleration \cite{khan2013exploratory} or to compute the arm orientation \cite{moazen2016airdraw}. The preprocessing phase is typically devoted to noise filtering and to the separation of gravity and body acceleration components. The latter procedure typically involves the use of a low-pass filter \cite{khan2013exploratory} \cite{bruno2014using}.

Acceleration data recorded during the execution of a gesture typically appear as a time series. In order to reduce the complexity of the classification problem some solutions suggest extracting discrete features using statistical analysis \cite{xie2016accelerometer}, the Haar Transform \cite{khan2012gesthaar} or extraction of the parameters from an autoregressive model \cite{khan2013exploratory}. The discrete features are then used to classify gestures using approaches based on Feed Forward Neural Networks (FNN) \cite{xie2016accelerometer} \cite{khan2013exploratory} or Support Vector Machines (SVMs) \cite{khan2012gesthaar}. An alternative approach envisions the use of time series to build time-dependent models, for example, using Gaussian Mixture Modeling (GMM) and Gaussian Mixture Regression (GMR) \cite{bruno2014using}, to extract continuous features such as the sensor orientation \cite{srivastava2016hand} or to simply store them as templates \cite{moazen2016airdraw} \cite{akl2010accelerometer} \cite{liu2009uwave} \cite{wu2010hand} \cite{porzi2013smart}, and then use techniques such as Dynamic Time Warping (DTW) \cite{moazen2016airdraw} \cite{wu2010hand} \cite{srivastava2016hand} to compare them with the data that should be classified. DTW is a \textit{de facto standard} solution in the literature, possibly combined with other methods such as affinity propagation \cite{akl2010accelerometer} and template adaptation \cite{liu2009uwave}. Adopted alternatives to DTW are represented by Mahalanobis distance \cite{bruno2014using}, Global Alignment Kernel \cite{porzi2013smart} and Recurrent Neural Network classifiers \cite{shin2016dynamic}. Moreover the possibility to classify human gestures using the prediction error generated by a Continuous Time Recurrent Neural Network (CTRNN) predictor has been explored  \cite{bailador2007real}.


Whenever the processing of the acceleration data is expected to be done online, the problem of recognising gestures should encompass their detection. The accelerometer time series should be segmented to isolate the portion of data where a gesture is detected. Simple segmentation approaches require the end-user to communicate through buttons \cite{xie2016accelerometer} \cite{akl2010accelerometer} \cite{liu2009uwave} or touch-screens \cite{porzi2013smart} when a gesture starts and ends. More advanced approaches typically focus on detecting variations in the data stream \cite{xie2016accelerometer} \cite{moazen2016airdraw} \cite{wu2010hand}. The segmentation induces a sporadic gesture recognition whose main limitation is that the gesture must necessarily finish before the classification process starts. Literature presents very few examples of gesture recognition systems able to perform online, continuous recognition. One solution proposes the use of a moving horizon window, for continuous gesture recognition, combined with a threshold mechanism to discriminate between unknown and known gestures \cite{bruno2014using}.

The objective of this paper is to investigate the integration of an RNN probabilistic classifier, whose performances have been assessed in \cite{shin2016dynamic}, in an architecture that uses a moving horizon window, as in \cite{bruno2014using}, to ensure a continuous, \textit{as-early-as-possible}, gesture recognition. Specifically, SLOTH uses raw inertial data and relies on a novel mechanism, which models gestures occurrences on top of neural network output patterns, for discriminating between known and unknown gestures.

\section{System's Architecture}
\label{sec:sofar}
\begin{figure}
	\centering
	\vspace{0.2 cm}
	\includegraphics[width=0.9\linewidth, trim={0cm 0cm 0cm 0cm},clip]{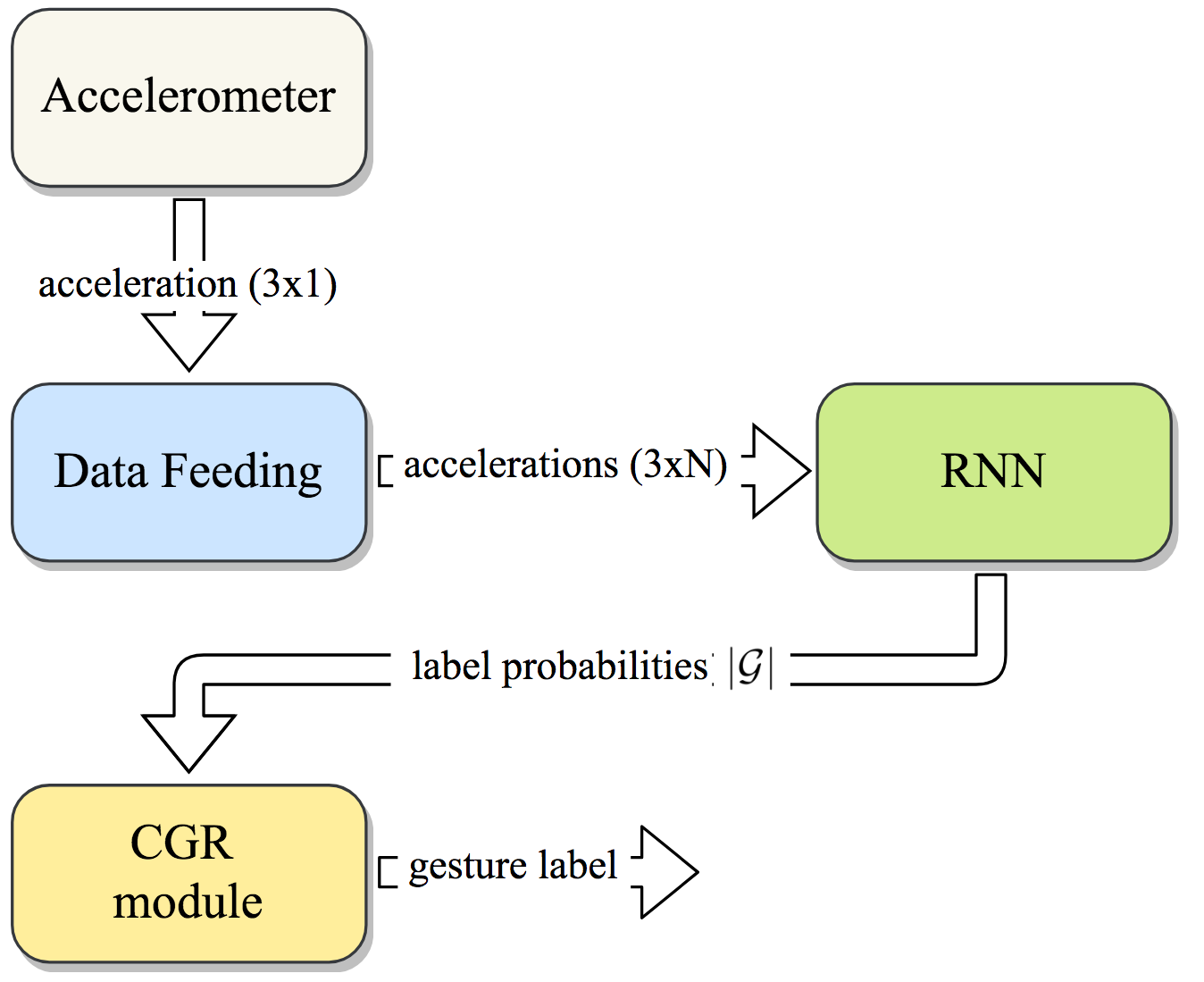}
	\caption{SLOTH's architecture.} 
	\label{fig:Architecture}
\end{figure}
\noindent
SLOTH processes data collected by a triaxial accelerometer worn by users on their right wrist and, whenever a gesture is recognised, it returns a label. As described in Fig. \ref{fig:Architecture}, the overall architecture is composed of three modules: \textit{Data Feeding}, \textit{Recurrent Neural Network} (\textit{RNN}) and ${\textit{Continuous Gesture Recognition}}$ (\textit{CGR}). 
\subsection{Data Feeding} 
\label{feeding}
\noindent
The \textit{Data Feeding} module receives raw acceleration data from a triaxial accelerometer at a fixed frequency $f$ and stores them in a buffer of size $N$, where $N$ depends on gestures length. Once the buffer is full (i.e., after $N$ time instants), the \textit{Data Feeding} module sends the content of the buffer to the RNN module. At each new sample, the content of the buffer is shifted forward to include the new sample and the updated buffer content is sent to the RNN. The \textit{Data Feeding} module does not introduce time steps delay.
\subsection{Recurrent Neural Network}
\label{RNN}
\begin{figure}
\centering
\vspace{0.2 cm}
\includegraphics[width=\linewidth, trim={1cm 0cm 0cm 0cm},clip]{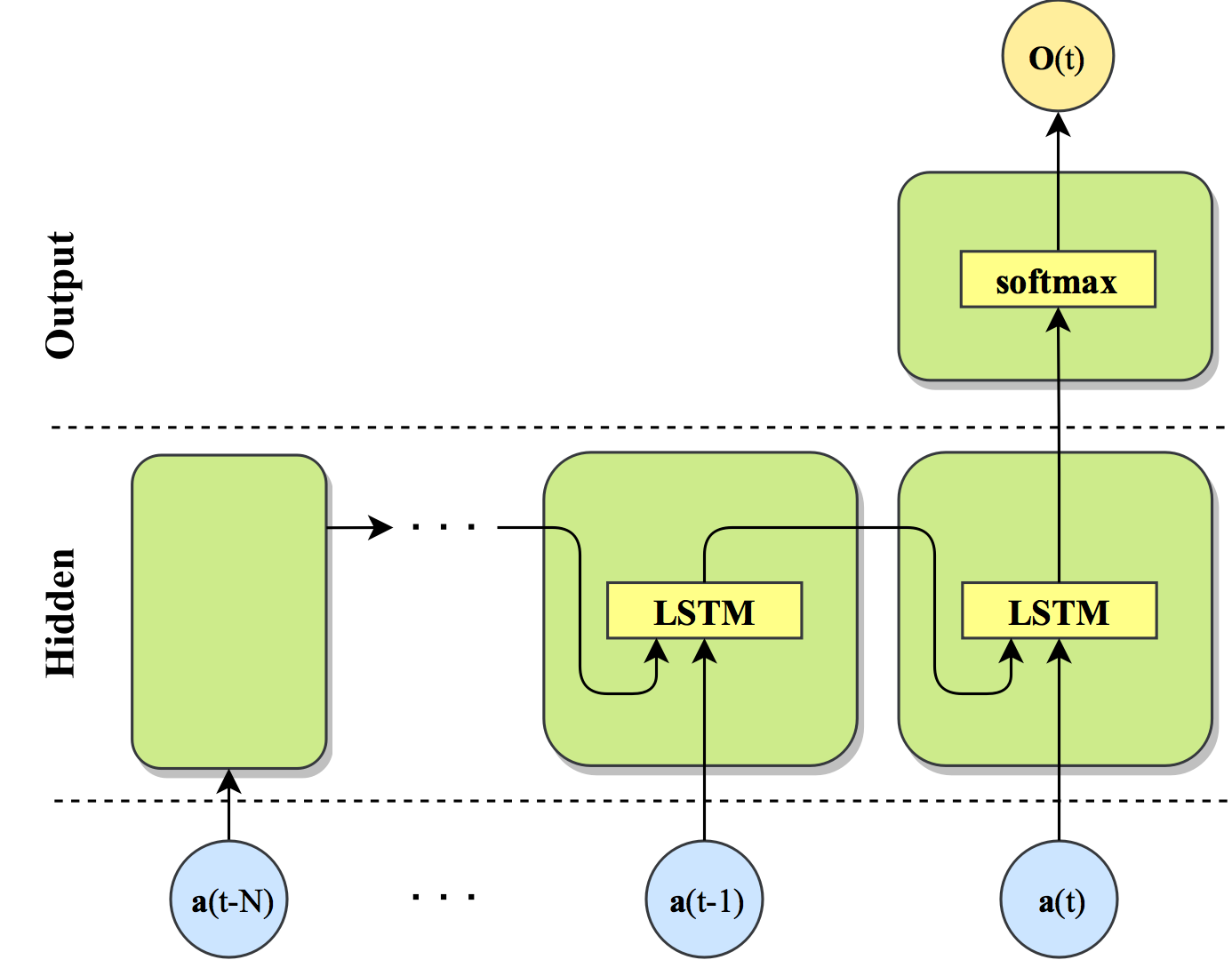}
\caption{The unfolded computational graph of the RNN. The RNN is composed of an LSTM recurrent layer to model temporal relations and a softmax layer for classification. The network receives as input a triaxial acceleration time series and outputs label confidences.}
\label{fig:RNN_structure}
\end{figure}
\noindent
An RNN, structured as in Fig. \ref{fig:RNN_structure}, has been chosen for its capability to model time-dependent behaviours to perform a probabilistic classification of gestures using time series of triaxial linear accelerations. The RNN receives as input a time series ${\mbox{\boldmath$a$}(t)=[a_x(t), a_y(t), a_z(t)]}$. The input is fed to a Long Short-Term Memory (LSTM) hidden layer, which learns long-term temporal dependencies. The output of the hidden layer, for the last input time step, is fed to a \textit{softmax} output layer that returns the probabilities for ${\mbox{\boldmath$a$}(t)}$ belonging to each considered gesture. The network, working under closed-world assumption, discriminates between gesture classes described in the dictionary:
\begin{equation}
	\mathcal{G}=\{G_1, \dots, G_{|\mathcal{G}|}\},
\end{equation}
Each gesture class $G_i$ is assumed to be \textit{unique} (i.e., a data stream cannot be classified as an instance of different gesture classes at the same time) and \textit{independent} (i.e., each class is not related to, as a component or sub-part of, other classes), and characterised by an 
average temporal duration $S_i$. As described in Fig. \ref{fig:Architecture} the RNN receives as input a time series $\mbox{\boldmath$a$}(t)$ of dimension $3\times N$ with 
\begin{equation}
	\label{eq:temporal_size}
	N = \max_{i=1, \dots, |\mathcal{G}|}(S_i)
\end{equation}
Since the network has been trained over $|\mathcal{G}|$ gesture classes, the output vector $\mbox{\boldmath$o$}(t)$ has dimension $|\mathcal{G}|$. During the training of the RNN, beside acceleration samples from $g_j \in G_j$, a target vector {\boldmath$v$} for $\mbox{\boldmath$o$}(t)$ is given, normalised such that all values are zero except for ${v_{G_j} = 1}$. Therefore, when the trained network is used, each element ${o_i \in [0, 1]}$ and, when acceleration data from $g_j$ are given as input to the network, $o_j$ tends to one while others tend to zero.
\subsection{Continuous Gesture Recognition}
\label{voting}
\begin{figure}
\centering
\includegraphics[width=\linewidth]{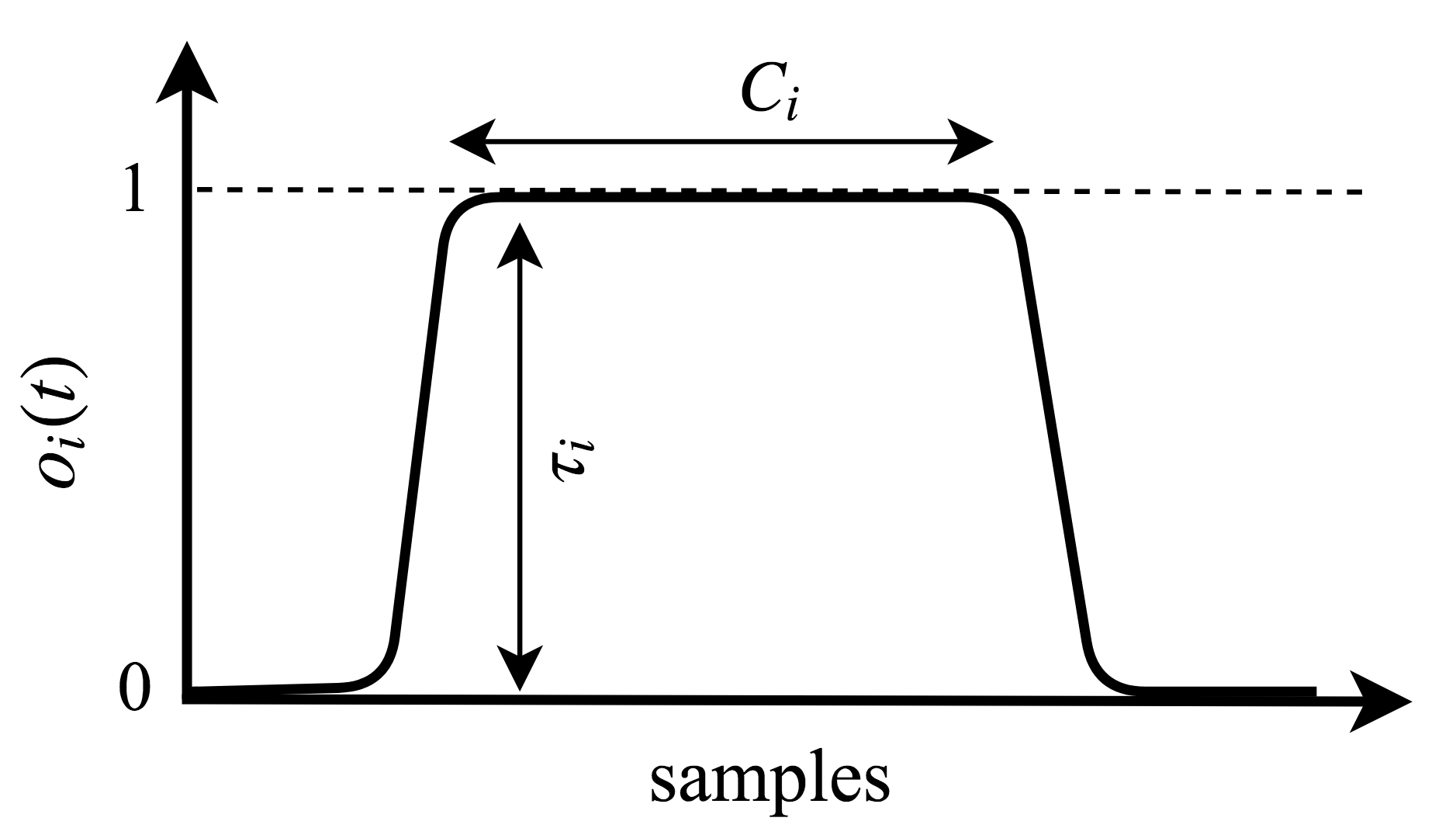}
\caption{Plateau behaviour expected in $o_i(t)$ when gesture $G_i$ occurs.}
\label{fig:Model}
\end{figure}
\noindent
The CGR module receives the neural network output $\mbox{\boldmath$o$}(t)$ representing probabilities associated with each gesture class. The GCR module processes the stream of gesture probabilities to detect and classify known gestures, relaxing the closed-word assumption introduced by the RNN (necessary condition to implement a continuous recognition). 

As described in Sec. \ref{RNN} the neural network reacts at time $t$ to gesture $g_i$ by raising $o_i(t)$ to 1. This implies a positive peak in the derivative:
\begin{equation}
	\Delta o_i(t) = o_i(t) - o_i(t-1).
\end{equation}
Since it is the derivative of $o_i(t)$,  ${\Delta o_i(t)}$ is a scalar of value in the interval ${[0, 1]}$. We define the peak instant $t_p$ for the gesture $g_i$ as the time instant for which ${\Delta o_i(t_p) > \rho}$ . The threshold $\rho$ allows for filtering out small fluctuations due to noise.

The network is trained with many $g_i$ examples that differ in time length and signal magnitude, therefore the resulting network is able to recognise temporal pattern associated with $G_i$ in different conditions. Furthermore, since the considered gestures are unique and independent (Sec. \ref{RNN}), their temporal patterns are unique and independent as well and the network needs to process only a portion of the gesture before being able to classify it. For these reasons and because of the buffering mechanism, the expected $o_i(t)$ behaviour when $g_i$ occurs is represented by a plateau as in Fig. \ref{fig:Model}.

Due to the buffering performed by the \textit{Data Feeding} module every sample ${\mbox{\boldmath$a$}(t)}$ with $t>N$ is processed $N$ times. Assuming as a classification limit case the presence in the buffer, as first or last element, one sample ${\mbox{\boldmath$a$}(t) \in g_i}$, the previously described model can be formalised as:
\begin{equation}
	\label{eq:limit}
	\lim_{\tau_i \to 1} \left(\tau_i - \frac{1}{C_i}\sum_{t_p}^{t_p+C_i}o_i(t) \right)=0^-,
\end{equation}
where
\begin{equation}
	\label{eq:condition}
	 \frac{1}{C_i}\sum_{t_p}^{t_p+C_i}o_i(t) = A_i \leq 1,
\end{equation}
and  ${C_i < S_i + N}$. In particular, \eqref{eq:limit} describes the plateau behaviour of $o_i(t)$, while \eqref{eq:condition} describes the limit case, when the network classified perfectly $G_i$ for $C_i$ samples, then ${A_i=1}$. The described model implies that when $g_i$ occurs, then ${A_i \geq \tau_i}$. Note that in \eqref{eq:limit} $\tau$ and $C$ are presented as gesture dependent parameters, in fact ideally the network response should be homogeneous for all the gesture classes but this does not typically happen, thus $\tau_i$ and $C_i$ should be defined experimentally.

Iteratively and independently for all the gesture classes in $\mathcal{G}$, the CGR module:
\begin{itemize}
	\item identifies positive peaks (\textit{detection});
	\item classifies the samples in the input buffer as an occurrence of gesture class $G_i$ if the condition ${A_i \geq \tau_i}$ is satisfied (\textit{classification}).
\end{itemize}
Different buffer shifts containing samples referring to a single $g_i$ could satisfy the condition ${A_i \geq \tau_i}$. Therefore, in order to avoid $g_i$ to be recognised multiple times, each positive peak is associated with only one recognition.
\section{Dataset}
\label{sec:dataset}
\begin{figure}
	\centering
	\vspace{0.2 cm}
	\includegraphics[width=\linewidth]{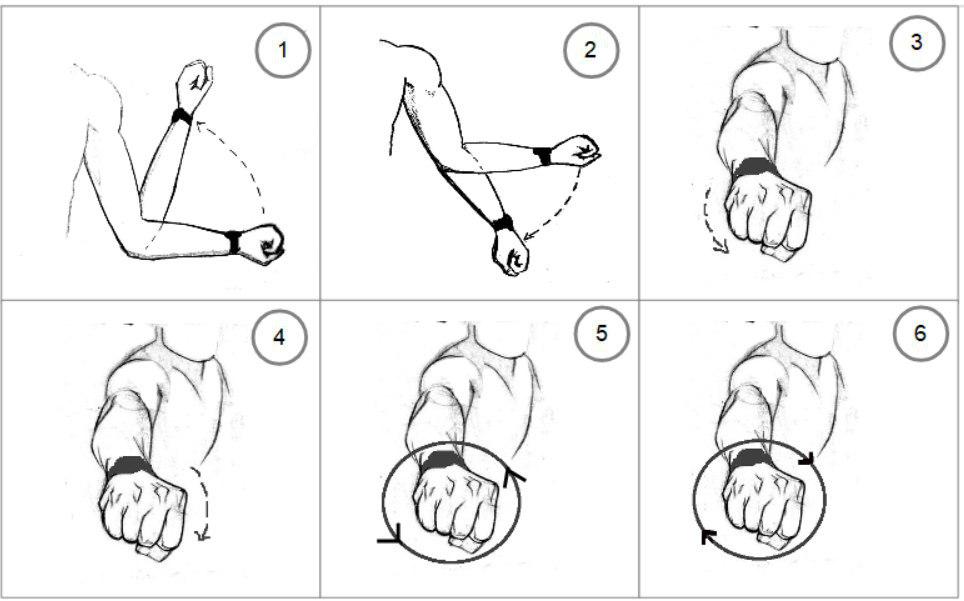}
	\caption{Visual representation of the gestures composing the dictionary.}
	\label{fig:gestures}
\end{figure}
\noindent
Experiments to acquire right wrist acceleration data are performed using an LG G Watch R smartwatch. The smartwatch is equipped with a triaxial accelerometer and it is paired with a smart-phone that receives the data and saves them on file. The system collects data at a frequency ${f=40 \hspace{0,1 cm}\textit{Hz}}$, this data are then downsampled at ${10 \hspace{0,1 cm}\textit{Hz}}$. The gesture dictionary is composed of the six gestures represented in Fig. \ref{fig:gestures}. All gestures assume the same starting pose for the arm: the elbow bent at 90 degrees while in contact with the flank and the hand held horizontally and pointing forward. Similarly, all gestures end when the arm is back in the starting pose. As described in Fig. \ref{fig:gestures}, in $G_1$ the arm moves upward maintaining fixed the elbow and with no wrist twist, in $G_2$ the arm moves downward maintaining fixed the elbow and with no wrist twist, in $G_3$ the arm is stretched and the wrist twists clockwise, in $G_4$ the arm is stretched and the wrist twists counter-clockwise, in $G_5$ the hand performs a clockwise circle with no wrist twist and, lastly, whereas in $G_6$ the hand performs an anti-clockwise circle with no wrist twist. Using the afore-described equipment, we collected two datasets of the six gestures composing the vocabulary, which we refer to as \textit{Dataset A} and \textit{Dataset B}:

\textit{Dataset A} is used to train and test the RNN module. Ten volunteers performed nine times the six gestures described above, providing a total of 540 sequences. These sequences are manually cut so that they only contain acceleration samples which refer to the execution of the gestures. The dataset has been divided, preserving the balance of volunteers and gestures, in two subsets, respectively used for the training ($70\%$) and testing ($30\%$) of the RNN module.

\textit{Dataset B} includes 15 sequences collected from one volunteer, known by the system. While in \textit{Dataset A} one sequence refers to one execution of one gesture, sequences in \textit{Dataset B} contains from a minimum of 6 to a maximum of 12 gestures (providing approximately 20 executions per gesture), separated by a non-constant number of samples in which the user remains in the starting pose. There are no consecutive executions of the same gesture in the sequences. The dataset is manually tagged. 

\section{Implementation}
\label{sec:implementation}
\noindent
The modelling and recognition system presented above has been implemented in MATLAB R2017b.

\textit{Data Feeding}. In the tests with sequences of \textit{ Dataset B}, the \textit{Data Feeding} module is in charge of simulating the online usage of the architecture. Given a sequence belonging to \textit{Dataset B}, it loads the acceleration data sample by sample and feeds them to the RNN module through the buffer, whose size has been set to ${N=40}$ samples.

\textit{Recurrent Neural Network}. 
\begin{figure}[!t]
	\centering
	\vspace{0.2 cm}
	\setlength\figureheight{0.8\linewidth}
	\setlength\figurewidth{0.8\linewidth}
	\footnotesize \input{./confusion_matrix_offline.tex}
	\caption{Confusion matrix for the RNN offline testing. The bottom row reports the \textit{recall} measures while the rightmost column reports the \textit{precision} measures. The blue cell reports the overall accuracy.}
	\label{fig:confusion_matrix}
\end{figure}
As shown in Fig. \ref{fig:RNN_structure}, the RNN is composed of an LSTM layer and a \textit{softmax} layer, which are implemented using standard MATLAB libraries. The hidden layer is composed of 32 neurons, and the training procedure uses the \textit{cross entropy} loss function and \textit{stochastic gradient descendent with momentum} as an optimiser. Since the results of \eqref{eq:temporal_size} for ${\textit{Dataset A}}$ is ${N=40}$, the input size of the network is ${3 \times 40}$. Therefore the training sequences $\mbox{\boldmath$a$}(t)$ containing less than 40 samples are padded with $\mbox{\boldmath$a$}(1)$ at the beginning. During the training and the offline testing phases, the buffer mechanism is not present and for each sequence $\mbox{\boldmath$a$}(t)$ the network returns a single vector {\boldmath$O$}. Since the selected gesture dictionary has dimension $\mathcal{G}=6$, the size of {\boldmath$O$} is $6$ as well. The network output {\boldmath$O$}$(k)$ for each sequence $k$ in the test set, containing $g_i$, is processed by an \textit{argmax} function to determine the $i$-label. Fig. \ref{fig:confusion_matrix} shows the confusion matrix obtained by the RNN on the testing \textit{Dataset A}. It can be seen that, the RNN achieves good results in terms of accuracy, precision and recall.

\textit{Continuous Gesture Recognition}. The GCR module has three parameters, {\boldmath$C$}, {\boldmath$\tau$} and $\rho$, which must be set according to the gesture dictionary and to the neural network response. In order to filter out only small fluctuations, in the interval of possible values ${[0, 1]}$, it is picked ${\rho=0.2}$. Instead, {\boldmath$C$} and {\boldmath$\tau$} can be defined as:
\begin{equation}
\label{eq:parameters}
\begin{aligned}
&\mbox{\boldmath$C$} = \alpha (\mbox{\boldmath$S$}+N), \hspace{1 cm} &0<\alpha\leq 1,\\
&\mbox{\boldmath$\tau$} = \gamma \mbox{\boldmath$M$}, &0<\gamma\leq 1,
\end{aligned}
\end{equation}
where {\boldmath$M$} is the average network response for each gesture, such that
\begin{equation}
	M_i = \frac{1}{n_i}\sum_{k=1}^{n_i}O_i(k)
\end{equation}
given that $n_i$ is the number of sequences contained in the dataset referring to $G_i$.
From an analysis of \textit{Dataset A}, it results:
\begin{equation}
\begin{aligned}
	&\mbox{\boldmath$S$} = [38, 38, 37, 37, 38, 38], \\
	&\mbox{\boldmath$M$} = [0.996, 0.996, 0.97, 0.995, 0.934, 0.917].
\end{aligned}
\end{equation}
Setting ${\alpha = 0.25}$ and ${\gamma = 0.9}$ leads to:
\begin{equation}
\label{eq:parameters}
\begin{aligned}
	&\rho=0.2,\\
	&\mbox{\boldmath$C$} = [20, 20, 19, 19, 20, 20],\\
	&\mbox{\boldmath$\tau$} = [0.896, 0.896, 0.873, 0.895, 0.84, 0.825]
\end{aligned}
\end{equation}
as final set of parameters.
\begin{figure}[!t]
	\centering
	\vspace{0.2 cm}
	\setlength\figureheight{0.8\linewidth}
	\setlength\figurewidth{0.8\linewidth}
	\footnotesize \input{./confusion_matrix_online.tex}
	\caption{Confusion matrix for the online testing with parameters from \eqref{eq:parameters}. The bottom row reports the \textit{recall} measures while the rightmost column reports the \textit{precision} measures. The blue cell reports the overall accuracy.}
	\label{fig:confusion_matrix_online}
\end{figure}
\section{Experimental Evaluation}
\label{sec:results}
\begin{figure}[t]
	\centering
	\vspace{0.2 cm}
	\setlength\figureheight{0.8\linewidth}
	\setlength\figurewidth{0.8\linewidth}
	\footnotesize \input{./confusion_matrix_online_05.tex}
	\caption{Confusion matrix for the online testing with ${\alpha = 0.05}$ and ${\gamma = 0.9}$. The bottom row reports the \textit{recall} measures while the rightmost column reports the \textit{precision} measures. The blue cell reports the overall accuracy.}
	\label{fig:confusion_matrix_5}
\end{figure}
\begin{figure}[t]
	\centering
	\vspace{0.2 cm}
	\setlength\figureheight{0.8\linewidth}
	\setlength\figurewidth{0.8\linewidth}
	\footnotesize \input{./confusion_matrix_online_60.tex}
	\caption{Confusion matrix for the online testing with ${\alpha = 0.25}$ and ${\gamma = 0.6}$. The bottom row reports the \textit{recall} measures while the rightmost column report the \textit{precision} measures. The blue cell reports the overall accuracy.}
	\label{fig:confusion_matrix_60}
\end{figure}
\begin{figure*}[!tbph]
	\centering
	\vspace{0.2 cm}
	\subfloat[Classification output for an online test with system parameters from \eqref{eq:parameters}.]{\includegraphics[width=.99\linewidth, trim={5.5cm 2.5cm 5.5cm 2cm},clip]{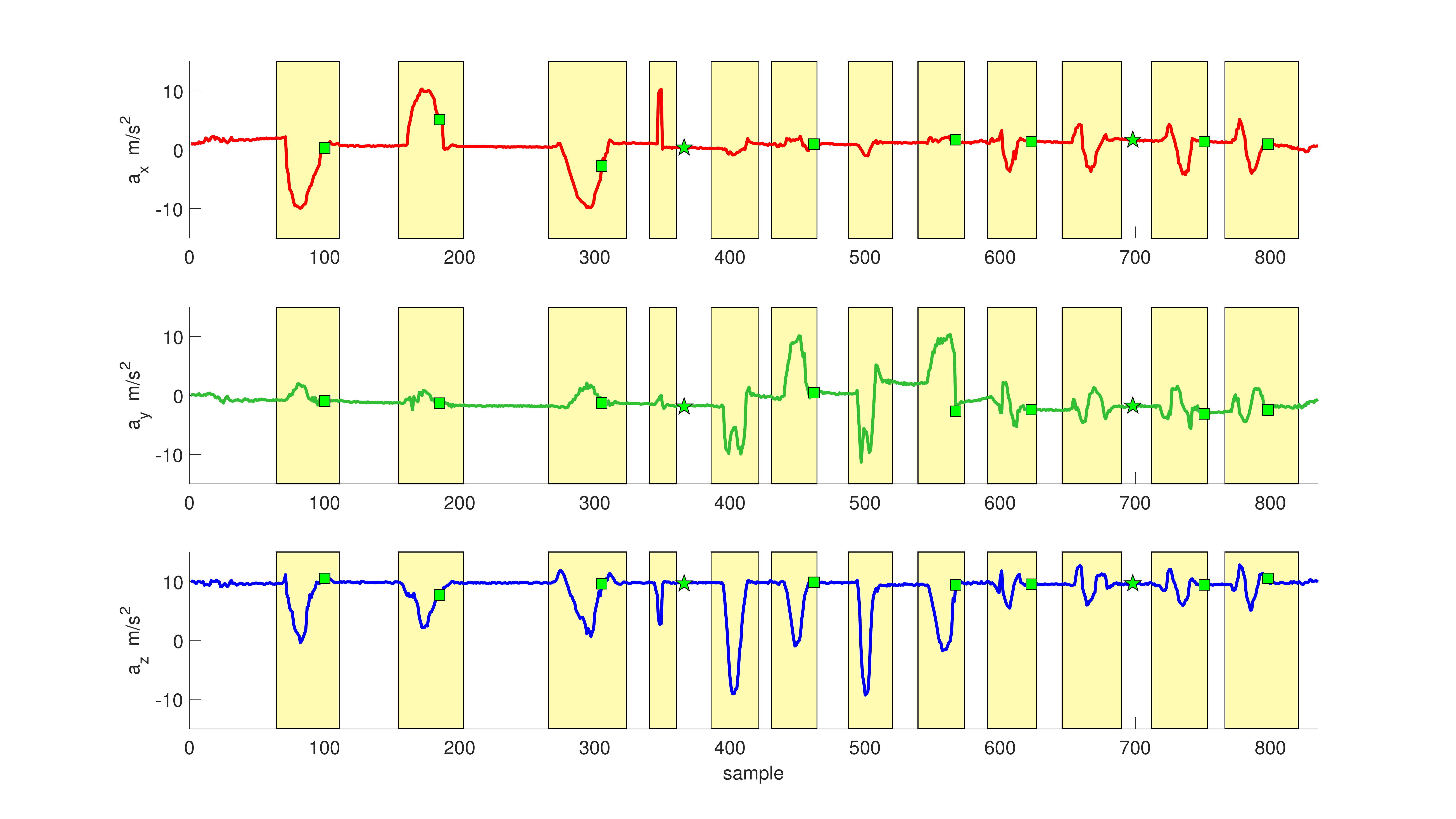} \label{fig:121234345656}}
	\vspace{0.5cm}
	\subfloat[Classification output for an online test with ${\alpha = 0.05}$ and ${\gamma = 0.9}$]{\includegraphics[width=.99\linewidth, trim={5.5cm 2.5cm 5.5cm 2cm},clip]{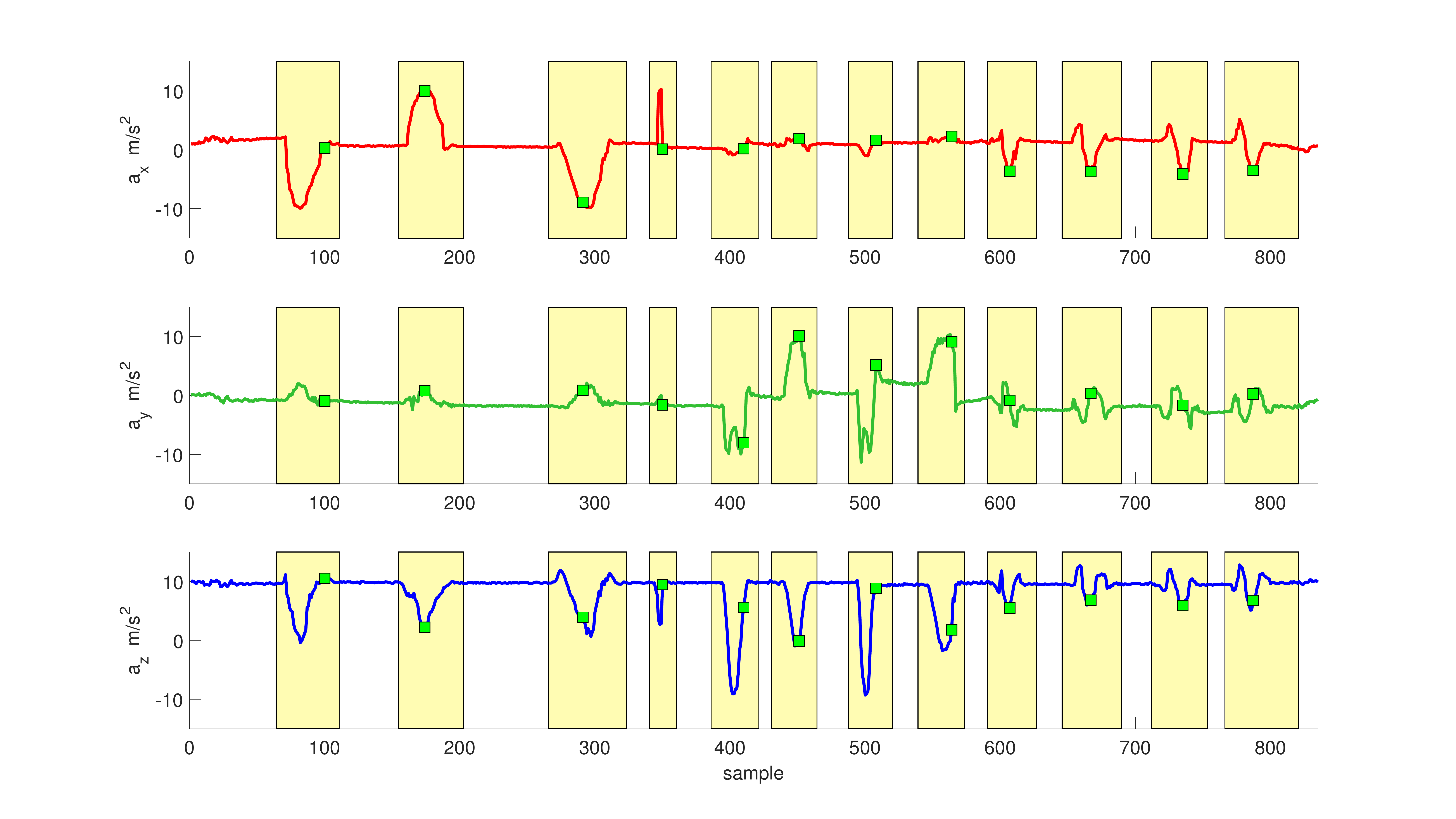} \label{fig:121234345656_5}}
	\caption{From top to bottom are represented the x, y and z acceleration components. Yellow boxes denote gestures instances, while green squares and stars denote correct classifications. Green squares denote when the classification occurred before the end of the gesture and green stars denote when the classification occurred after the end of the gesture.}
\end{figure*}

\noindent
Fig. \ref{fig:confusion_matrix_online} shows the confusion matrix obtained by testing SLOTH, in the implementation presented above, with the sequences of \textit{Dataset B}. The CGR module presented in Sec. \ref{voting} allows for relaxing the closed-word assumption, which is represented in Fig. \ref{fig:confusion_matrix_online} using the tag ``N. G.'' (Not a Gesture). In the figure, it is possible to observe that, the precision is very high for all gestures (the minimum is $94.4\%$ for $G_5$), while the recall is lower, especially for gestures $G_3$ ($55\%$) and $G_6$ ($45.5\%$). In both cases, most of the misclassified executions are not recognised at all (N.G.). Fig. \ref{fig:121234345656} presents the recognition results and the timings for one continuous sequence included in \textit{Dataset B} which contains each gesture twice (specifically, in the order ${G_1, G_2, G_1, G_2, G_3, G_4, G_3, G_4, G_5, G_6, G_5, G_6}$). The three graphs in Fig. \ref{fig:121234345656} show, from top to bottom, the $x$, $y$ and $z$ acceleration components. Yellow boxes denote gesture instances, while green squares and stars denote correct classifications. More precisely, green squares indicate when the recognition occurs before the end of the gesture while green stars denote when the recognition occurs after the end of the gesture. As Fig. \ref{fig:121234345656} shows, out of the 12 gestures contained in that recording, 10 are correctly classified and before their end, 2 are correctly classified after their end and 2 are not classified. 

The tests on \textit{Dataset B} reported in Fig. \ref{fig:confusion_matrix} and Fig. \ref{fig:121234345656}, show that the parameter settings discussed in Sec. \ref{sec:implementation} are very conservative, giving a clear preference to precision over recall. This behaviour is well suited for applications where gestures are used to control a robot, for example, but it may not be desirable in other contexts. Parameters {\boldmath$\tau$} and {\boldmath$C$} allow for controlling this behaviour. In particular, increasing these values makes the expected plateau longer ({\boldmath$C$}) and higher ({\boldmath$\tau$}), thus increasing precision, while reducing them makes the expected plateau shorter ({\boldmath$C$}) and lower ({\boldmath$\tau$}), thus increasing the recall. Furthermore reducing {\boldmath$C$} allows for recognising gestures earlier, thereby increasing the reactivity of the system. To verify whether and to what extent the above statement holds, we have repeated the tests on \textit{Dataset B} two more times, once decreasing {\boldmath$C$} to ${\alpha=0.05}$ while keeping {\boldmath$\tau$} to the value defined in \eqref{eq:parameters}, and one decreasing {\boldmath$\tau$} to ${\gamma=0.6}$ while keeping {\boldmath$C$} as defined in \eqref{eq:parameters}. The results of the first test are shown in Fig. \ref{fig:confusion_matrix_5} and Fig. \ref{fig:121234345656_5}, while the results of the second test are shown in Fig. \ref{fig:confusion_matrix_60}.

Fig. \ref{fig:confusion_matrix_5} shows that, as expected, new {\boldmath$C$} values yield an increase in the recall at the expenses of a small decrease in precision. Moreover, the number of samples required to issue the label (see Fig. \ref{fig:121234345656_5}) is significantly smaller than that with the values defined in \eqref{eq:parameters}. Similarly, Fig. \ref{fig:confusion_matrix_60} shows that new {\boldmath$\tau$} values yield an increase in the recall at the expenses of a small decrease in precision.

All the performed tests as well as the RNN offline testing presented in Fig. \ref{fig:confusion_matrix} highlight a difficulty in classifying of $G_6$. This is probably  a consequence of using raw acceleration data, which include a component related to gravity and one, in our case, related to the person's arm movements. When a person performs gestures $G_1$, $G_2$, $G_3$ or $G_4$, the gravity component shifts from one accelerometer axis to another, thus ensuring that the acceleration patterns encode sensible variations. This does not happen in the case of gestures $G_5$ and $G_6$. Since the gravity acceleration is by far the most prominent acceleration component, we argue that its shift between accelerometer axes helps the classification and, as a consequence, its absence causes the performance loss.

It is worth noticing that even in the configurations prioritising recall over precision, precision remains very high, thus proving the robustness of the proposed approach.

\section{Conclusions}
\label{sec:conclusions}
\noindent
We propose SLOTH an architecture for continuous human gesture recognition based on an LSTM Recurrent Neural Network probabilistic classifier, and a continuous gesture recognition module that does not require a segmentation procedure. The procedure relies on two parameters, {\boldmath$\tau$} and {\boldmath$C$}, to tune the recognition and prioritise precision over early recognition, or vice-versa. 

SLOTH has been tested with six hand gestures, over a dataset composed of $15$ gesture sequences. Experiments performed using different combinations for the CGR module parameters show that the proposed online gesture recognition system achieves on average  very good precision, up to $99\%$, and recall, up to $97\%$.  The main drawback of our approach is that the RNN needs to be retrained every time a gesture is added/deleted, and therefore the system's performance depends on the chosen combination of gestures. Future developments of this work will include an extensive study of how {\boldmath$C$} and {\boldmath$\tau$} affect the classification performance, an online implementation and a comparison study with state-of-the-art methods in terms of performance and classification time. Furthermore, it will be explored the possibility to integrate SLOTH in architectures for gesture-based robot control \cite{coronado2017gesture} and human-robot cooperation \cite{darvish2018flexible}.

\addtolength{\textheight}{-12cm}   



\bibliographystyle{ieeeconf}
\bibliography{bibliography}

\end{document}

%% file: confusion_matrix_offline.tex
%
%
\definecolor{mycolor1}{rgb}{0.47059,0.90196,0.70588}%
\definecolor{mycolor2}{rgb}{0.90196,0.54902,0.54902}%
\definecolor{mycolor3}{rgb}{0.47059,0.58824,0.90196}%
\begin{tikzpicture}

\begin{axis}[%
width=\figureheight,
height=\figurewidth,
at={(4.795in,0.693in)},
scale only axis,
xmin=0.5,
xmax=7.5,
xtick={1,2,3,4,5,6,7,8},
xticklabels={{$\text{G}_\text{1}$},{$\text{G}_\text{2}$},{$\text{G}_\text{3}$},{$\text{G}_\text{4}$},{$\text{G}_\text{5}$},{$\text{G}_\text{6}$},{},{1}},
xlabel style={font=\color{white!15!black}},
xlabel={Target Class},
y dir=reverse,
ymin=0.5,
ymax=7.5,
ytick={1,2,3,4,5,6,7,8},
yticklabels={{$\text{G}_\text{1}$},{$\text{G}_\text{2}$},{$\text{G}_\text{3}$},{$\text{G}_\text{4}$},{$\text{G}_\text{5}$},{$\text{G}_\text{6}$},{},{1}},
ylabel style={font=\color{white!15!black}},
ylabel={Output Class},
axis background/.style={fill=white},
title style={font=\bfseries},
title={ Confusion Matrix},
legend style={legend cell align=left, align=left, draw=white!15!black}
]

\addplot[area legend, draw=black, fill=mycolor1]
table[row sep=crcr] {%
x	y\\
0.5	0.5\\
1.5	0.5\\
1.5	1.5\\
0.5	1.5\\
}--cycle;

\node[above, align=center, font=\bfseries]
at (axis cs:1,1) {26};
\node[below, align=center]
at (axis cs:1,1) {16.0\%};

\addplot[area legend, draw=black, fill=mycolor2]
table[row sep=crcr] {%
x	y\\
0.5	1.5\\
1.5	1.5\\
1.5	2.5\\
0.5	2.5\\
}--cycle;

\node[above, align=center, font=\bfseries]
at (axis cs:1,2) {0};
\node[below, align=center]
at (axis cs:1,2) {0.0\%};

\addplot[area legend, draw=black, fill=mycolor2]
table[row sep=crcr] {%
x	y\\
0.5	2.5\\
1.5	2.5\\
1.5	3.5\\
0.5	3.5\\
}--cycle;

\node[above, align=center, font=\bfseries]
at (axis cs:1,3) {0};
\node[below, align=center]
at (axis cs:1,3) {0.0\%};

\addplot[area legend, draw=black, fill=mycolor2]
table[row sep=crcr] {%
x	y\\
0.5	3.5\\
1.5	3.5\\
1.5	4.5\\
0.5	4.5\\
}--cycle;

\node[above, align=center, font=\bfseries]
at (axis cs:1,4) {0};
\node[below, align=center]
at (axis cs:1,4) {0.0\%};

\addplot[area legend, draw=black, fill=mycolor2]
table[row sep=crcr] {%
x	y\\
0.5	4.5\\
1.5	4.5\\
1.5	5.5\\
0.5	5.5\\
}--cycle;

\node[above, align=center, font=\bfseries]
at (axis cs:1,5) {0};
\node[below, align=center]
at (axis cs:1,5) {0.0\%};

\addplot[area legend, draw=black, fill=mycolor2]
table[row sep=crcr] {%
x	y\\
0.5	5.5\\
1.5	5.5\\
1.5	6.5\\
0.5	6.5\\
}--cycle;

\node[above, align=center, font=\bfseries]
at (axis cs:1,6) {1};
\node[below, align=center]
at (axis cs:1,6) {0.6\%};

\addplot[area legend, draw=black, fill=gray]
table[row sep=crcr] {%
x	y\\
0.5	6.5\\
1.5	6.5\\
1.5	7.5\\
0.5	7.5\\
}--cycle;

\node[above, align=center, font=\color{black!60!green}]
at (axis cs:1,7) {96.3\%};
\node[below, align=center, font=\color{black!60!red}]
at (axis cs:1,7) {3.7\%};

\addplot[area legend, draw=black, fill=mycolor2]
table[row sep=crcr] {%
x	y\\
1.5	0.5\\
2.5	0.5\\
2.5	1.5\\
1.5	1.5\\
}--cycle;

\node[above, align=center, font=\bfseries]
at (axis cs:2,1) {0};
\node[below, align=center]
at (axis cs:2,1) {0.0\%};

\addplot[area legend, draw=black, fill=mycolor1]
table[row sep=crcr] {%
x	y\\
1.5	1.5\\
2.5	1.5\\
2.5	2.5\\
1.5	2.5\\
}--cycle;

\node[above, align=center, font=\bfseries]
at (axis cs:2,2) {27};
\node[below, align=center]
at (axis cs:2,2) {16.7\%};

\addplot[area legend, draw=black, fill=mycolor2]
table[row sep=crcr] {%
x	y\\
1.5	2.5\\
2.5	2.5\\
2.5	3.5\\
1.5	3.5\\
}--cycle;

\node[above, align=center, font=\bfseries]
at (axis cs:2,3) {0};
\node[below, align=center]
at (axis cs:2,3) {0.0\%};

\addplot[area legend, draw=black, fill=mycolor2]
table[row sep=crcr] {%
x	y\\
1.5	3.5\\
2.5	3.5\\
2.5	4.5\\
1.5	4.5\\
}--cycle;

\node[above, align=center, font=\bfseries]
at (axis cs:2,4) {0};
\node[below, align=center]
at (axis cs:2,4) {0.0\%};

\addplot[area legend, draw=black, fill=mycolor2]
table[row sep=crcr] {%
x	y\\
1.5	4.5\\
2.5	4.5\\
2.5	5.5\\
1.5	5.5\\
}--cycle;

\node[above, align=center, font=\bfseries]
at (axis cs:2,5) {0};
\node[below, align=center]
at (axis cs:2,5) {0.0\%};

\addplot[area legend, draw=black, fill=mycolor2]
table[row sep=crcr] {%
x	y\\
1.5	5.5\\
2.5	5.5\\
2.5	6.5\\
1.5	6.5\\
}--cycle;

\node[above, align=center, font=\bfseries]
at (axis cs:2,6) {0};
\node[below, align=center]
at (axis cs:2,6) {0.0\%};

\addplot[area legend, draw=black, fill=gray]
table[row sep=crcr] {%
x	y\\
1.5	6.5\\
2.5	6.5\\
2.5	7.5\\
1.5	7.5\\
}--cycle;

\node[above, align=center, font=\color{black!60!green}]
at (axis cs:2,7) {100\%};
\node[below, align=center, font=\color{black!60!red}]
at (axis cs:2,7) {0.0\%};

\addplot[area legend, draw=black, fill=mycolor2]
table[row sep=crcr] {%
x	y\\
2.5	0.5\\
3.5	0.5\\
3.5	1.5\\
2.5	1.5\\
}--cycle;

\node[above, align=center, font=\bfseries]
at (axis cs:3,1) {0};
\node[below, align=center]
at (axis cs:3,1) {0.0\%};

\addplot[area legend, draw=black, fill=mycolor2]
table[row sep=crcr] {%
x	y\\
2.5	1.5\\
3.5	1.5\\
3.5	2.5\\
2.5	2.5\\
}--cycle;

\node[above, align=center, font=\bfseries]
at (axis cs:3,2) {0};
\node[below, align=center]
at (axis cs:3,2) {0.0\%};

\addplot[area legend, draw=black, fill=mycolor1]
table[row sep=crcr] {%
x	y\\
2.5	2.5\\
3.5	2.5\\
3.5	3.5\\
2.5	3.5\\
}--cycle;

\node[above, align=center, font=\bfseries]
at (axis cs:3,3) {27};
\node[below, align=center]
at (axis cs:3,3) {16.7\%};

\addplot[area legend, draw=black, fill=mycolor2]
table[row sep=crcr] {%
x	y\\
2.5	3.5\\
3.5	3.5\\
3.5	4.5\\
2.5	4.5\\
}--cycle;

\node[above, align=center, font=\bfseries]
at (axis cs:3,4) {0};
\node[below, align=center]
at (axis cs:3,4) {0.0\%};

\addplot[area legend, draw=black, fill=mycolor2]
table[row sep=crcr] {%
x	y\\
2.5	4.5\\
3.5	4.5\\
3.5	5.5\\
2.5	5.5\\
}--cycle;

\node[above, align=center, font=\bfseries]
at (axis cs:3,5) {0};
\node[below, align=center]
at (axis cs:3,5) {0.0\%};

\addplot[area legend, draw=black, fill=mycolor2]
table[row sep=crcr] {%
x	y\\
2.5	5.5\\
3.5	5.5\\
3.5	6.5\\
2.5	6.5\\
}--cycle;

\node[above, align=center, font=\bfseries]
at (axis cs:3,6) {0};
\node[below, align=center]
at (axis cs:3,6) {0.0\%};

\addplot[area legend, draw=black, fill=gray]
table[row sep=crcr] {%
x	y\\
2.5	6.5\\
3.5	6.5\\
3.5	7.5\\
2.5	7.5\\
}--cycle;

\node[above, align=center, font=\color{black!60!green}]
at (axis cs:3,7) {100\%};
\node[below, align=center, font=\color{black!60!red}]
at (axis cs:3,7) {0.0\%};

\addplot[area legend, draw=black, fill=mycolor2]
table[row sep=crcr] {%
x	y\\
3.5	0.5\\
4.5	0.5\\
4.5	1.5\\
3.5	1.5\\
}--cycle;

\node[above, align=center, font=\bfseries]
at (axis cs:4,1) {0};
\node[below, align=center]
at (axis cs:4,1) {0.0\%};

\addplot[area legend, draw=black, fill=mycolor2]
table[row sep=crcr] {%
x	y\\
3.5	1.5\\
4.5	1.5\\
4.5	2.5\\
3.5	2.5\\
}--cycle;

\node[above, align=center, font=\bfseries]
at (axis cs:4,2) {0};
\node[below, align=center]
at (axis cs:4,2) {0.0\%};

\addplot[area legend, draw=black, fill=mycolor2]
table[row sep=crcr] {%
x	y\\
3.5	2.5\\
4.5	2.5\\
4.5	3.5\\
3.5	3.5\\
}--cycle;

\node[above, align=center, font=\bfseries]
at (axis cs:4,3) {0};
\node[below, align=center]
at (axis cs:4,3) {0.0\%};

\addplot[area legend, draw=black, fill=mycolor1]
table[row sep=crcr] {%
x	y\\
3.5	3.5\\
4.5	3.5\\
4.5	4.5\\
3.5	4.5\\
}--cycle;

\node[above, align=center, font=\bfseries]
at (axis cs:4,4) {27};
\node[below, align=center]
at (axis cs:4,4) {16.7\%};

\addplot[area legend, draw=black, fill=mycolor2]
table[row sep=crcr] {%
x	y\\
3.5	4.5\\
4.5	4.5\\
4.5	5.5\\
3.5	5.5\\
}--cycle;

\node[above, align=center, font=\bfseries]
at (axis cs:4,5) {0};
\node[below, align=center]
at (axis cs:4,5) {0.0\%};

\addplot[area legend, draw=black, fill=mycolor2]
table[row sep=crcr] {%
x	y\\
3.5	5.5\\
4.5	5.5\\
4.5	6.5\\
3.5	6.5\\
}--cycle;

\node[above, align=center, font=\bfseries]
at (axis cs:4,6) {0};
\node[below, align=center]
at (axis cs:4,6) {0.0\%};

\addplot[area legend, draw=black, fill=gray]
table[row sep=crcr] {%
x	y\\
3.5	6.5\\
4.5	6.5\\
4.5	7.5\\
3.5	7.5\\
}--cycle;

\node[above, align=center, font=\color{black!60!green}]
at (axis cs:4,7) {100\%};
\node[below, align=center, font=\color{black!60!red}]
at (axis cs:4,7) {0.0\%};

\addplot[area legend, draw=black, fill=mycolor2]
table[row sep=crcr] {%
x	y\\
4.5	0.5\\
5.5	0.5\\
5.5	1.5\\
4.5	1.5\\
}--cycle;

\node[above, align=center, font=\bfseries]
at (axis cs:5,1) {0};
\node[below, align=center]
at (axis cs:5,1) {0.0\%};

\addplot[area legend, draw=black, fill=mycolor2]
table[row sep=crcr] {%
x	y\\
4.5	1.5\\
5.5	1.5\\
5.5	2.5\\
4.5	2.5\\
}--cycle;

\node[above, align=center, font=\bfseries]
at (axis cs:5,2) {0};
\node[below, align=center]
at (axis cs:5,2) {0.0\%};

\addplot[area legend, draw=black, fill=mycolor2]
table[row sep=crcr] {%
x	y\\
4.5	2.5\\
5.5	2.5\\
5.5	3.5\\
4.5	3.5\\
}--cycle;

\node[above, align=center, font=\bfseries]
at (axis cs:5,3) {0};
\node[below, align=center]
at (axis cs:5,3) {0.0\%};

\addplot[area legend, draw=black, fill=mycolor2]
table[row sep=crcr] {%
x	y\\
4.5	3.5\\
5.5	3.5\\
5.5	4.5\\
4.5	4.5\\
}--cycle;

\node[above, align=center, font=\bfseries]
at (axis cs:5,4) {0};
\node[below, align=center]
at (axis cs:5,4) {0.0\%};

\addplot[area legend, draw=black, fill=mycolor1]
table[row sep=crcr] {%
x	y\\
4.5	4.5\\
5.5	4.5\\
5.5	5.5\\
4.5	5.5\\
}--cycle;

\node[above, align=center, font=\bfseries]
at (axis cs:5,5) {27};
\node[below, align=center]
at (axis cs:5,5) {16.7\%};

\addplot[area legend, draw=black, fill=mycolor2]
table[row sep=crcr] {%
x	y\\
4.5	5.5\\
5.5	5.5\\
5.5	6.5\\
4.5	6.5\\
}--cycle;

\node[above, align=center, font=\bfseries]
at (axis cs:5,6) {0};
\node[below, align=center]
at (axis cs:5,6) {0.0\%};

\addplot[area legend, draw=black, fill=gray]
table[row sep=crcr] {%
x	y\\
4.5	6.5\\
5.5	6.5\\
5.5	7.5\\
4.5	7.5\\
}--cycle;

\node[above, align=center, font=\color{black!60!green}]
at (axis cs:5,7) {100\%};
\node[below, align=center, font=\color{black!60!red}]
at (axis cs:5,7) {0.0\%};

\addplot[area legend, draw=black, fill=mycolor2]
table[row sep=crcr] {%
x	y\\
5.5	0.5\\
6.5	0.5\\
6.5	1.5\\
5.5	1.5\\
}--cycle;

\node[above, align=center, font=\bfseries]
at (axis cs:6,1) {0};
\node[below, align=center]
at (axis cs:6,1) {0.0\%};

\addplot[area legend, draw=black, fill=mycolor2]
table[row sep=crcr] {%
x	y\\
5.5	1.5\\
6.5	1.5\\
6.5	2.5\\
5.5	2.5\\
}--cycle;

\node[above, align=center, font=\bfseries]
at (axis cs:6,2) {1};
\node[below, align=center]
at (axis cs:6,2) {0.6\%};

\addplot[area legend, draw=black, fill=mycolor2]
table[row sep=crcr] {%
x	y\\
5.5	2.5\\
6.5	2.5\\
6.5	3.5\\
5.5	3.5\\
}--cycle;

\node[above, align=center, font=\bfseries]
at (axis cs:6,3) {0};
\node[below, align=center]
at (axis cs:6,3) {0.0\%};

\addplot[area legend, draw=black, fill=mycolor2]
table[row sep=crcr] {%
x	y\\
5.5	3.5\\
6.5	3.5\\
6.5	4.5\\
5.5	4.5\\
}--cycle;

\node[above, align=center, font=\bfseries]
at (axis cs:6,4) {0};
\node[below, align=center]
at (axis cs:6,4) {0.0\%};

\addplot[area legend, draw=black, fill=mycolor2]
table[row sep=crcr] {%
x	y\\
5.5	4.5\\
6.5	4.5\\
6.5	5.5\\
5.5	5.5\\
}--cycle;

\node[above, align=center, font=\bfseries]
at (axis cs:6,5) {3};
\node[below, align=center]
at (axis cs:6,5) {1.9\%};

\addplot[area legend, draw=black, fill=mycolor1]
table[row sep=crcr] {%
x	y\\
5.5	5.5\\
6.5	5.5\\
6.5	6.5\\
5.5	6.5\\
}--cycle;

\node[above, align=center, font=\bfseries]
at (axis cs:6,6) {23};
\node[below, align=center]
at (axis cs:6,6) {14.2\%};

\addplot[area legend, draw=black, fill=gray]
table[row sep=crcr] {%
x	y\\
5.5	6.5\\
6.5	6.5\\
6.5	7.5\\
5.5	7.5\\
}--cycle;

\node[above, align=center, font=\color{black!60!green}]
at (axis cs:6,7) {85.2\%};
\node[below, align=center, font=\color{black!60!red}]
at (axis cs:6,7) {14.8\%};

\addplot[area legend, draw=black, fill=gray]
table[row sep=crcr] {%
x	y\\
6.5	0.5\\
7.5	0.5\\
7.5	1.5\\
6.5	1.5\\
}--cycle;

\node[above, align=center, font=\color{black!60!green}]
at (axis cs:7,1) {100\%};
\node[below, align=center, font=\color{black!60!red}]
at (axis cs:7,1) {0.0\%};

\addplot[area legend, draw=black, fill=gray]
table[row sep=crcr] {%
x	y\\
6.5	1.5\\
7.5	1.5\\
7.5	2.5\\
6.5	2.5\\
}--cycle;

\node[above, align=center, font=\color{black!60!green}]
at (axis cs:7,2) {96.4\%};
\node[below, align=center, font=\color{black!60!red}]
at (axis cs:7,2) {3.6\%};

\addplot[area legend, draw=black, fill=gray]
table[row sep=crcr] {%
x	y\\
6.5	2.5\\
7.5	2.5\\
7.5	3.5\\
6.5	3.5\\
}--cycle;

\node[above, align=center, font=\color{black!60!green}]
at (axis cs:7,3) {100\%};
\node[below, align=center, font=\color{black!60!red}]
at (axis cs:7,3) {0.0\%};

\addplot[area legend, draw=black, fill=gray]
table[row sep=crcr] {%
x	y\\
6.5	3.5\\
7.5	3.5\\
7.5	4.5\\
6.5	4.5\\
}--cycle;

\node[above, align=center, font=\color{black!60!green}]
at (axis cs:7,4) {100\%};
\node[below, align=center, font=\color{black!60!red}]
at (axis cs:7,4) {0.0\%};

\addplot[area legend, draw=black, fill=gray]
table[row sep=crcr] {%
x	y\\
6.5	4.5\\
7.5	4.5\\
7.5	5.5\\
6.5	5.5\\
}--cycle;

\node[above, align=center, font=\color{black!60!green}]
at (axis cs:7,5) {90.0\%};
\node[below, align=center, font=\color{black!60!red}]
at (axis cs:7,5) {10.0\%};

\addplot[area legend, draw=black, fill=gray]
table[row sep=crcr] {%
x	y\\
6.5	5.5\\
7.5	5.5\\
7.5	6.5\\
6.5	6.5\\
}--cycle;

\node[above, align=center, font=\color{black!60!green}]
at (axis cs:7,6) {95.8\%};
\node[below, align=center, font=\color{black!60!red}]
at (axis cs:7,6) {4.2\%};

\addplot[area legend, draw=black, fill=mycolor3]
table[row sep=crcr] {%
x	y\\
6.5	6.5\\
7.5	6.5\\
7.5	7.5\\
6.5	7.5\\
}--cycle;

\node[above, align=center, font=\bfseries\color{black!60!green}]
at (axis cs:7,7) {96.9\%};
\node[below, align=center, font=\bfseries\color{black!60!red}]
at (axis cs:7,7) {3.1\%};
\addplot [color=darkgray, line width=2.0pt]
  table[row sep=crcr]{%
6.5	0.5\\
6.5	7.5\\
};

\addplot [color=darkgray, line width=2.0pt, forget plot]
  table[row sep=crcr]{%
0.5	6.5\\
7.5	6.5\\
};
\end{axis}
\end{tikzpicture}%

%% file: confusion_matrix_online.tex
%
%
\definecolor{mycolor1}{rgb}{0.47059,0.90196,0.70588}%
\definecolor{mycolor2}{rgb}{0.90196,0.54902,0.54902}%
\definecolor{mycolor3}{rgb}{0.47059,0.58824,0.90196}%
\begin{tikzpicture}

\begin{axis}[%
width=\figureheight,
height=\figureheight,
at={(4.795in,0.693in)},
scale only axis,
xmin=0.5,
xmax=8.5,
xtick={1,2,3,4,5,6,7,8,9},
xticklabels={{$\text{G}_\text{1}$},{$\text{G}_\text{2}$},{$\text{G}_\text{3}$},{$\text{G}_\text{4}$},{$\text{G}_\text{5}$},{$\text{G}_\text{6}$},{N.G.},{},{1}},
xlabel style={font=\color{white!15!black}},
xlabel={Target Class},
y dir=reverse,
ymin=0.5,
ymax=8.5,
ytick={1,2,3,4,5,6,7,8,9},
yticklabels={{$\text{G}_\text{1}$},{$\text{G}_\text{2}$},{$\text{G}_\text{3}$},{$\text{G}_\text{4}$},{$\text{G}_\text{5}$},{$\text{G}_\text{6}$},{N.G.},{},{1}},
ylabel style={font=\color{white!15!black}},
ylabel={Output Class},
axis background/.style={fill=white},
title style={font=\bfseries},
title={ Confusion Matrix},
legend style={legend cell align=left, align=left, draw=white!15!black}
]

\addplot[area legend, draw=black, fill=mycolor1]
table[row sep=crcr] {%
x	y\\
0.5	0.5\\
1.5	0.5\\
1.5	1.5\\
0.5	1.5\\
}--cycle;

\node[above, align=center, font=\bfseries]
at (axis cs:1,1) {20};
\node[below, align=center]
at (axis cs:1,1) {15.6\%};

\addplot[area legend, draw=black, fill=mycolor2]
table[row sep=crcr] {%
x	y\\
0.5	1.5\\
1.5	1.5\\
1.5	2.5\\
0.5	2.5\\
}--cycle;

\node[above, align=center, font=\bfseries]
at (axis cs:1,2) {0};
\node[below, align=center]
at (axis cs:1,2) {0.0\%};

\addplot[area legend, draw=black, fill=mycolor2]
table[row sep=crcr] {%
x	y\\
0.5	2.5\\
1.5	2.5\\
1.5	3.5\\
0.5	3.5\\
}--cycle;

\node[above, align=center, font=\bfseries]
at (axis cs:1,3) {0};
\node[below, align=center]
at (axis cs:1,3) {0.0\%};

\addplot[area legend, draw=black, fill=mycolor2]
table[row sep=crcr] {%
x	y\\
0.5	3.5\\
1.5	3.5\\
1.5	4.5\\
0.5	4.5\\
}--cycle;

\node[above, align=center, font=\bfseries]
at (axis cs:1,4) {0};
\node[below, align=center]
at (axis cs:1,4) {0.0\%};

\addplot[area legend, draw=black, fill=mycolor2]
table[row sep=crcr] {%
x	y\\
0.5	4.5\\
1.5	4.5\\
1.5	5.5\\
0.5	5.5\\
}--cycle;

\node[above, align=center, font=\bfseries]
at (axis cs:1,5) {0};
\node[below, align=center]
at (axis cs:1,5) {0.0\%};

\addplot[area legend, draw=black, fill=mycolor2]
table[row sep=crcr] {%
x	y\\
0.5	5.5\\
1.5	5.5\\
1.5	6.5\\
0.5	6.5\\
}--cycle;

\node[above, align=center, font=\bfseries]
at (axis cs:1,6) {0};
\node[below, align=center]
at (axis cs:1,6) {0.0\%};

\addplot[area legend, draw=black, fill=mycolor2]
table[row sep=crcr] {%
x	y\\
0.5	6.5\\
1.5	6.5\\
1.5	7.5\\
0.5	7.5\\
}--cycle;

\node[above, align=center, font=\bfseries]
at (axis cs:1,7) {0};
\node[below, align=center]
at (axis cs:1,7) {0.0\%};

\addplot[area legend, draw=black, fill=gray]
table[row sep=crcr] {%
x	y\\
0.5	7.5\\
1.5	7.5\\
1.5	8.5\\
0.5	8.5\\
}--cycle;

\node[above, align=center, font=\color{black!60!green}]
at (axis cs:1,8) {100\%};
\node[below, align=center, font=\color{black!60!red}]
at (axis cs:1,8) {0.0\%};

\addplot[area legend, draw=black, fill=mycolor2]
table[row sep=crcr] {%
x	y\\
1.5	0.5\\
2.5	0.5\\
2.5	1.5\\
1.5	1.5\\
}--cycle;

\node[above, align=center, font=\bfseries]
at (axis cs:2,1) {0};
\node[below, align=center]
at (axis cs:2,1) {0.0\%};

\addplot[area legend, draw=black, fill=mycolor1]
table[row sep=crcr] {%
x	y\\
1.5	1.5\\
2.5	1.5\\
2.5	2.5\\
1.5	2.5\\
}--cycle;

\node[above, align=center, font=\bfseries]
at (axis cs:2,2) {22};
\node[below, align=center]
at (axis cs:2,2) {17.2\%};

\addplot[area legend, draw=black, fill=mycolor2]
table[row sep=crcr] {%
x	y\\
1.5	2.5\\
2.5	2.5\\
2.5	3.5\\
1.5	3.5\\
}--cycle;

\node[above, align=center, font=\bfseries]
at (axis cs:2,3) {0};
\node[below, align=center]
at (axis cs:2,3) {0.0\%};

\addplot[area legend, draw=black, fill=mycolor2]
table[row sep=crcr] {%
x	y\\
1.5	3.5\\
2.5	3.5\\
2.5	4.5\\
1.5	4.5\\
}--cycle;

\node[above, align=center, font=\bfseries]
at (axis cs:2,4) {0};
\node[below, align=center]
at (axis cs:2,4) {0.0\%};

\addplot[area legend, draw=black, fill=mycolor2]
table[row sep=crcr] {%
x	y\\
1.5	4.5\\
2.5	4.5\\
2.5	5.5\\
1.5	5.5\\
}--cycle;

\node[above, align=center, font=\bfseries]
at (axis cs:2,5) {0};
\node[below, align=center]
at (axis cs:2,5) {0.0\%};

\addplot[area legend, draw=black, fill=mycolor2]
table[row sep=crcr] {%
x	y\\
1.5	5.5\\
2.5	5.5\\
2.5	6.5\\
1.5	6.5\\
}--cycle;

\node[above, align=center, font=\bfseries]
at (axis cs:2,6) {0};
\node[below, align=center]
at (axis cs:2,6) {0.0\%};

\addplot[area legend, draw=black, fill=mycolor2]
table[row sep=crcr] {%
x	y\\
1.5	6.5\\
2.5	6.5\\
2.5	7.5\\
1.5	7.5\\
}--cycle;

\node[above, align=center, font=\bfseries]
at (axis cs:2,7) {0};
\node[below, align=center]
at (axis cs:2,7) {0.0\%};

\addplot[area legend, draw=black, fill=gray]
table[row sep=crcr] {%
x	y\\
1.5	7.5\\
2.5	7.5\\
2.5	8.5\\
1.5	8.5\\
}--cycle;

\node[above, align=center, font=\color{black!60!green}]
at (axis cs:2,8) {100\%};
\node[below, align=center, font=\color{black!60!red}]
at (axis cs:2,8) {0.0\%};

\addplot[area legend, draw=black, fill=mycolor2]
table[row sep=crcr] {%
x	y\\
2.5	0.5\\
3.5	0.5\\
3.5	1.5\\
2.5	1.5\\
}--cycle;

\node[above, align=center, font=\bfseries]
at (axis cs:3,1) {0};
\node[below, align=center]
at (axis cs:3,1) {0.0\%};

\addplot[area legend, draw=black, fill=mycolor2]
table[row sep=crcr] {%
x	y\\
2.5	1.5\\
3.5	1.5\\
3.5	2.5\\
2.5	2.5\\
}--cycle;

\node[above, align=center, font=\bfseries]
at (axis cs:3,2) {0};
\node[below, align=center]
at (axis cs:3,2) {0.0\%};

\addplot[area legend, draw=black, fill=mycolor1]
table[row sep=crcr] {%
x	y\\
2.5	2.5\\
3.5	2.5\\
3.5	3.5\\
2.5	3.5\\
}--cycle;

\node[above, align=center, font=\bfseries]
at (axis cs:3,3) {11};
\node[below, align=center]
at (axis cs:3,3) {8.6\%};

\addplot[area legend, draw=black, fill=mycolor2]
table[row sep=crcr] {%
x	y\\
2.5	3.5\\
3.5	3.5\\
3.5	4.5\\
2.5	4.5\\
}--cycle;

\node[above, align=center, font=\bfseries]
at (axis cs:3,4) {0};
\node[below, align=center]
at (axis cs:3,4) {0.0\%};

\addplot[area legend, draw=black, fill=mycolor2]
table[row sep=crcr] {%
x	y\\
2.5	4.5\\
3.5	4.5\\
3.5	5.5\\
2.5	5.5\\
}--cycle;

\node[above, align=center, font=\bfseries]
at (axis cs:3,5) {0};
\node[below, align=center]
at (axis cs:3,5) {0.0\%};

\addplot[area legend, draw=black, fill=mycolor2]
table[row sep=crcr] {%
x	y\\
2.5	5.5\\
3.5	5.5\\
3.5	6.5\\
2.5	6.5\\
}--cycle;

\node[above, align=center, font=\bfseries]
at (axis cs:3,6) {0};
\node[below, align=center]
at (axis cs:3,6) {0.0\%};

\addplot[area legend, draw=black, fill=mycolor2]
table[row sep=crcr] {%
x	y\\
2.5	6.5\\
3.5	6.5\\
3.5	7.5\\
2.5	7.5\\
}--cycle;

\node[above, align=center, font=\bfseries]
at (axis cs:3,7) {9};
\node[below, align=center]
at (axis cs:3,7) {7.0\%};

\addplot[area legend, draw=black, fill=gray]
table[row sep=crcr] {%
x	y\\
2.5	7.5\\
3.5	7.5\\
3.5	8.5\\
2.5	8.5\\
}--cycle;

\node[above, align=center, font=\color{black!60!green}]
at (axis cs:3,8) {55.0\%};
\node[below, align=center, font=\color{black!60!red}]
at (axis cs:3,8) {45.0\%};

\addplot[area legend, draw=black, fill=mycolor2]
table[row sep=crcr] {%
x	y\\
3.5	0.5\\
4.5	0.5\\
4.5	1.5\\
3.5	1.5\\
}--cycle;

\node[above, align=center, font=\bfseries]
at (axis cs:4,1) {0};
\node[below, align=center]
at (axis cs:4,1) {0.0\%};

\addplot[area legend, draw=black, fill=mycolor2]
table[row sep=crcr] {%
x	y\\
3.5	1.5\\
4.5	1.5\\
4.5	2.5\\
3.5	2.5\\
}--cycle;

\node[above, align=center, font=\bfseries]
at (axis cs:4,2) {0};
\node[below, align=center]
at (axis cs:4,2) {0.0\%};

\addplot[area legend, draw=black, fill=mycolor2]
table[row sep=crcr] {%
x	y\\
3.5	2.5\\
4.5	2.5\\
4.5	3.5\\
3.5	3.5\\
}--cycle;

\node[above, align=center, font=\bfseries]
at (axis cs:4,3) {0};
\node[below, align=center]
at (axis cs:4,3) {0.0\%};

\addplot[area legend, draw=black, fill=mycolor1]
table[row sep=crcr] {%
x	y\\
3.5	3.5\\
4.5	3.5\\
4.5	4.5\\
3.5	4.5\\
}--cycle;

\node[above, align=center, font=\bfseries]
at (axis cs:4,4) {22};
\node[below, align=center]
at (axis cs:4,4) {17.2\%};

\addplot[area legend, draw=black, fill=mycolor2]
table[row sep=crcr] {%
x	y\\
3.5	4.5\\
4.5	4.5\\
4.5	5.5\\
3.5	5.5\\
}--cycle;

\node[above, align=center, font=\bfseries]
at (axis cs:4,5) {0};
\node[below, align=center]
at (axis cs:4,5) {0.0\%};

\addplot[area legend, draw=black, fill=mycolor2]
table[row sep=crcr] {%
x	y\\
3.5	5.5\\
4.5	5.5\\
4.5	6.5\\
3.5	6.5\\
}--cycle;

\node[above, align=center, font=\bfseries]
at (axis cs:4,6) {0};
\node[below, align=center]
at (axis cs:4,6) {0.0\%};

\addplot[area legend, draw=black, fill=mycolor2]
table[row sep=crcr] {%
x	y\\
3.5	6.5\\
4.5	6.5\\
4.5	7.5\\
3.5	7.5\\
}--cycle;

\node[above, align=center, font=\bfseries]
at (axis cs:4,7) {1};
\node[below, align=center]
at (axis cs:4,7) {0.8\%};

\addplot[area legend, draw=black, fill=gray]
table[row sep=crcr] {%
x	y\\
3.5	7.5\\
4.5	7.5\\
4.5	8.5\\
3.5	8.5\\
}--cycle;

\node[above, align=center, font=\color{black!60!green}]
at (axis cs:4,8) {95.7\%};
\node[below, align=center, font=\color{black!60!red}]
at (axis cs:4,8) {4.3\%};

\addplot[area legend, draw=black, fill=mycolor2]
table[row sep=crcr] {%
x	y\\
4.5	0.5\\
5.5	0.5\\
5.5	1.5\\
4.5	1.5\\
}--cycle;

\node[above, align=center, font=\bfseries]
at (axis cs:5,1) {0};
\node[below, align=center]
at (axis cs:5,1) {0.0\%};

\addplot[area legend, draw=black, fill=mycolor2]
table[row sep=crcr] {%
x	y\\
4.5	1.5\\
5.5	1.5\\
5.5	2.5\\
4.5	2.5\\
}--cycle;

\node[above, align=center, font=\bfseries]
at (axis cs:5,2) {0};
\node[below, align=center]
at (axis cs:5,2) {0.0\%};

\addplot[area legend, draw=black, fill=mycolor2]
table[row sep=crcr] {%
x	y\\
4.5	2.5\\
5.5	2.5\\
5.5	3.5\\
4.5	3.5\\
}--cycle;

\node[above, align=center, font=\bfseries]
at (axis cs:5,3) {0};
\node[below, align=center]
at (axis cs:5,3) {0.0\%};

\addplot[area legend, draw=black, fill=mycolor2]
table[row sep=crcr] {%
x	y\\
4.5	3.5\\
5.5	3.5\\
5.5	4.5\\
4.5	4.5\\
}--cycle;

\node[above, align=center, font=\bfseries]
at (axis cs:5,4) {0};
\node[below, align=center]
at (axis cs:5,4) {0.0\%};

\addplot[area legend, draw=black, fill=mycolor1]
table[row sep=crcr] {%
x	y\\
4.5	4.5\\
5.5	4.5\\
5.5	5.5\\
4.5	5.5\\
}--cycle;

\node[above, align=center, font=\bfseries]
at (axis cs:5,5) {17};
\node[below, align=center]
at (axis cs:5,5) {13.3\%};

\addplot[area legend, draw=black, fill=mycolor2]
table[row sep=crcr] {%
x	y\\
4.5	5.5\\
5.5	5.5\\
5.5	6.5\\
4.5	6.5\\
}--cycle;

\node[above, align=center, font=\bfseries]
at (axis cs:5,6) {0};
\node[below, align=center]
at (axis cs:5,6) {0.0\%};

\addplot[area legend, draw=black, fill=mycolor2]
table[row sep=crcr] {%
x	y\\
4.5	6.5\\
5.5	6.5\\
5.5	7.5\\
4.5	7.5\\
}--cycle;

\node[above, align=center, font=\bfseries]
at (axis cs:5,7) {4};
\node[below, align=center]
at (axis cs:5,7) {3.1\%};

\addplot[area legend, draw=black, fill=gray]
table[row sep=crcr] {%
x	y\\
4.5	7.5\\
5.5	7.5\\
5.5	8.5\\
4.5	8.5\\
}--cycle;

\node[above, align=center, font=\color{black!60!green}]
at (axis cs:5,8) {81.0\%};
\node[below, align=center, font=\color{black!60!red}]
at (axis cs:5,8) {19.0\%};

\addplot[area legend, draw=black, fill=mycolor2]
table[row sep=crcr] {%
x	y\\
5.5	0.5\\
6.5	0.5\\
6.5	1.5\\
5.5	1.5\\
}--cycle;

\node[above, align=center, font=\bfseries]
at (axis cs:6,1) {0};
\node[below, align=center]
at (axis cs:6,1) {0.0\%};

\addplot[area legend, draw=black, fill=mycolor2]
table[row sep=crcr] {%
x	y\\
5.5	1.5\\
6.5	1.5\\
6.5	2.5\\
5.5	2.5\\
}--cycle;

\node[above, align=center, font=\bfseries]
at (axis cs:6,2) {0};
\node[below, align=center]
at (axis cs:6,2) {0.0\%};

\addplot[area legend, draw=black, fill=mycolor2]
table[row sep=crcr] {%
x	y\\
5.5	2.5\\
6.5	2.5\\
6.5	3.5\\
5.5	3.5\\
}--cycle;

\node[above, align=center, font=\bfseries]
at (axis cs:6,3) {0};
\node[below, align=center]
at (axis cs:6,3) {0.0\%};

\addplot[area legend, draw=black, fill=mycolor2]
table[row sep=crcr] {%
x	y\\
5.5	3.5\\
6.5	3.5\\
6.5	4.5\\
5.5	4.5\\
}--cycle;

\node[above, align=center, font=\bfseries]
at (axis cs:6,4) {0};
\node[below, align=center]
at (axis cs:6,4) {0.0\%};

\addplot[area legend, draw=black, fill=mycolor2]
table[row sep=crcr] {%
x	y\\
5.5	4.5\\
6.5	4.5\\
6.5	5.5\\
5.5	5.5\\
}--cycle;

\node[above, align=center, font=\bfseries]
at (axis cs:6,5) {1};
\node[below, align=center]
at (axis cs:6,5) {0.8\%};

\addplot[area legend, draw=black, fill=mycolor1]
table[row sep=crcr] {%
x	y\\
5.5	5.5\\
6.5	5.5\\
6.5	6.5\\
5.5	6.5\\
}--cycle;

\node[above, align=center, font=\bfseries]
at (axis cs:6,6) {10};
\node[below, align=center]
at (axis cs:6,6) {7.8\%};

\addplot[area legend, draw=black, fill=mycolor2]
table[row sep=crcr] {%
x	y\\
5.5	6.5\\
6.5	6.5\\
6.5	7.5\\
5.5	7.5\\
}--cycle;

\node[above, align=center, font=\bfseries]
at (axis cs:6,7) {11};
\node[below, align=center]
at (axis cs:6,7) {8.6\%};

\addplot[area legend, draw=black, fill=gray]
table[row sep=crcr] {%
x	y\\
5.5	7.5\\
6.5	7.5\\
6.5	8.5\\
5.5	8.5\\
}--cycle;

\node[above, align=center, font=\color{black!60!green}]
at (axis cs:6,8) {45.5\%};
\node[below, align=center, font=\color{black!60!red}]
at (axis cs:6,8) {54.5\%};

\addplot[area legend, draw=black, fill=mycolor2]
table[row sep=crcr] {%
x	y\\
6.5	0.5\\
7.5	0.5\\
7.5	1.5\\
6.5	1.5\\
}--cycle;

\node[above, align=center, font=\bfseries]
at (axis cs:7,1) {0};
\node[below, align=center]
at (axis cs:7,1) {0.0\%};

\addplot[area legend, draw=black, fill=mycolor2]
table[row sep=crcr] {%
x	y\\
6.5	1.5\\
7.5	1.5\\
7.5	2.5\\
6.5	2.5\\
}--cycle;

\node[above, align=center, font=\bfseries]
at (axis cs:7,2) {0};
\node[below, align=center]
at (axis cs:7,2) {0.0\%};

\addplot[area legend, draw=black, fill=mycolor2, forget plot]
table[row sep=crcr] {%
x	y\\
6.5	2.5\\
7.5	2.5\\
7.5	3.5\\
6.5	3.5\\
}--cycle;
\node[above, align=center, font=\bfseries]
at (axis cs:7,3) {0};
\node[below, align=center]
at (axis cs:7,3) {0.0\%};

\addplot[area legend, draw=black, fill=mycolor2, forget plot]
table[row sep=crcr] {%
x	y\\
6.5	3.5\\
7.5	3.5\\
7.5	4.5\\
6.5	4.5\\
}--cycle;
\node[above, align=center, font=\bfseries]
at (axis cs:7,4) {0};
\node[below, align=center]
at (axis cs:7,4) {0.0\%};

\addplot[area legend, draw=black, fill=mycolor2, forget plot]
table[row sep=crcr] {%
x	y\\
6.5	4.5\\
7.5	4.5\\
7.5	5.5\\
6.5	5.5\\
}--cycle;
\node[above, align=center, font=\bfseries]
at (axis cs:7,5) {0};
\node[below, align=center]
at (axis cs:7,5) {0.0\%};

\addplot[area legend, draw=black, fill=mycolor2, forget plot]
table[row sep=crcr] {%
x	y\\
6.5	5.5\\
7.5	5.5\\
7.5	6.5\\
6.5	6.5\\
}--cycle;
\node[above, align=center, font=\bfseries]
at (axis cs:7,6) {0};
\node[below, align=center]
at (axis cs:7,6) {0.0\%};

\addplot[area legend, draw=black, fill=mycolor1, forget plot]
table[row sep=crcr] {%
x	y\\
6.5	6.5\\
7.5	6.5\\
7.5	7.5\\
6.5	7.5\\
}--cycle;
\node[above, align=center, font=\bfseries]
at (axis cs:7,7) {0};
\node[below, align=center]
at (axis cs:7,7) {0.0\%};

\addplot[area legend, draw=black, fill=gray, forget plot]
table[row sep=crcr] {%
x	y\\
6.5	7.5\\
7.5	7.5\\
7.5	8.5\\
6.5	8.5\\
}--cycle;
\node[above, align=center, font=\color{black!60!green}]
at (axis cs:7,8) {NaN\%};
\node[below, align=center, font=\color{black!60!red}]
at (axis cs:7,8) {NaN\%};

\addplot[area legend, draw=black, fill=gray, forget plot]
table[row sep=crcr] {%
x	y\\
7.5	0.5\\
8.5	0.5\\
8.5	1.5\\
7.5	1.5\\
}--cycle;
\node[above, align=center, font=\color{black!60!green}]
at (axis cs:8,1) {100\%};
\node[below, align=center, font=\color{black!60!red}]
at (axis cs:8,1) {0.0\%};

\addplot[area legend, draw=black, fill=gray, forget plot]
table[row sep=crcr] {%
x	y\\
7.5	1.5\\
8.5	1.5\\
8.5	2.5\\
7.5	2.5\\
}--cycle;
\node[above, align=center, font=\color{black!60!green}]
at (axis cs:8,2) {100\%};
\node[below, align=center, font=\color{black!60!red}]
at (axis cs:8,2) {0.0\%};

\addplot[area legend, draw=black, fill=gray, forget plot]
table[row sep=crcr] {%
x	y\\
7.5	2.5\\
8.5	2.5\\
8.5	3.5\\
7.5	3.5\\
}--cycle;
\node[above, align=center, font=\color{black!60!green}]
at (axis cs:8,3) {100\%};
\node[below, align=center, font=\color{black!60!red}]
at (axis cs:8,3) {0.0\%};

\addplot[area legend, draw=black, fill=gray, forget plot]
table[row sep=crcr] {%
x	y\\
7.5	3.5\\
8.5	3.5\\
8.5	4.5\\
7.5	4.5\\
}--cycle;
\node[above, align=center, font=\color{black!60!green}]
at (axis cs:8,4) {100\%};
\node[below, align=center, font=\color{black!60!red}]
at (axis cs:8,4) {0.0\%};

\addplot[area legend, draw=black, fill=gray, forget plot]
table[row sep=crcr] {%
x	y\\
7.5	4.5\\
8.5	4.5\\
8.5	5.5\\
7.5	5.5\\
}--cycle;
\node[above, align=center, font=\color{black!60!green}]
at (axis cs:8,5) {94.4\%};
\node[below, align=center, font=\color{black!60!red}]
at (axis cs:8,5) {5.6\%};

\addplot[area legend, draw=black, fill=gray, forget plot]
table[row sep=crcr] {%
x	y\\
7.5	5.5\\
8.5	5.5\\
8.5	6.5\\
7.5	6.5\\
}--cycle;
\node[above, align=center, font=\color{black!60!green}]
at (axis cs:8,6) {100\%};
\node[below, align=center, font=\color{black!60!red}]
at (axis cs:8,6) {0.0\%};

\addplot[area legend, draw=black, fill=gray, forget plot]
table[row sep=crcr] {%
x	y\\
7.5	6.5\\
8.5	6.5\\
8.5	7.5\\
7.5	7.5\\
}--cycle;
\node[above, align=center, font=\color{black!60!green}]
at (axis cs:8,7) {0.0\%};
\node[below, align=center, font=\color{black!60!red}]
at (axis cs:8,7) {100\%};

\addplot[area legend, draw=black, fill=mycolor3, forget plot]
table[row sep=crcr] {%
x	y\\
7.5	7.5\\
8.5	7.5\\
8.5	8.5\\
7.5	8.5\\
}--cycle;
\node[above, align=center, font=\bfseries\color{black!60!green}]
at (axis cs:8,8) {79.7\%};
\node[below, align=center, font=\bfseries\color{black!60!red}]
at (axis cs:8,8) {20.3\%};
\addplot [color=darkgray, line width=2.0pt, forget plot]
  table[row sep=crcr]{%
7.5	0.5\\
7.5	8.5\\
};
\addplot [color=darkgray, line width=2.0pt, forget plot]
  table[row sep=crcr]{%
0.5	7.5\\
8.5	7.5\\
};
\end{axis}
\end{tikzpicture}%

%% file: confusion_matrix_online_05.tex
%
%
\definecolor{mycolor1}{rgb}{0.47059,0.90196,0.70588}%
\definecolor{mycolor2}{rgb}{0.90196,0.54902,0.54902}%
\definecolor{mycolor3}{rgb}{0.47059,0.58824,0.90196}%
\begin{tikzpicture}

\begin{axis}[%
width=\figureheight,
height=\figureheight,
at={(4.795in,0.693in)},
scale only axis,
xmin=0.5,
xmax=8.5,
xtick={1,2,3,4,5,6,7,8,9},
xticklabels={{$\text{G}_\text{1}$},{$\text{G}_\text{2}$},{$\text{G}_\text{3}$},{$\text{G}_\text{4}$},{$\text{G}_\text{5}$},{$\text{G}_\text{6}$},{N.G.},{},{1}},
xlabel style={font=\color{white!15!black}},
xlabel={Target Class},
y dir=reverse,
ymin=0.5,
ymax=8.5,
ytick={1,2,3,4,5,6,7,8,9},
yticklabels={{$\text{G}_\text{1}$},{$\text{G}_\text{2}$},{$\text{G}_\text{3}$},{$\text{G}_\text{4}$},{$\text{G}_\text{5}$},{$\text{G}_\text{6}$},{N.G.},{},{1}},
ylabel style={font=\color{white!15!black}},
ylabel={Output Class},
axis background/.style={fill=white},
title style={font=\bfseries},
title={ Confusion Matrix},
legend style={legend cell align=left, align=left, draw=white!15!black}
]

\addplot[area legend, draw=black, fill=mycolor1]
table[row sep=crcr] {%
x	y\\
0.5	0.5\\
1.5	0.5\\
1.5	1.5\\
0.5	1.5\\
}--cycle;

\node[above, align=center, font=\bfseries]
at (axis cs:1,1) {20};
\node[below, align=center]
at (axis cs:1,1) {14.8\%};

\addplot[area legend, draw=black, fill=mycolor2]
table[row sep=crcr] {%
x	y\\
0.5	1.5\\
1.5	1.5\\
1.5	2.5\\
0.5	2.5\\
}--cycle;

\node[above, align=center, font=\bfseries]
at (axis cs:1,2) {0};
\node[below, align=center]
at (axis cs:1,2) {0.0\%};

\addplot[area legend, draw=black, fill=mycolor2]
table[row sep=crcr] {%
x	y\\
0.5	2.5\\
1.5	2.5\\
1.5	3.5\\
0.5	3.5\\
}--cycle;

\node[above, align=center, font=\bfseries]
at (axis cs:1,3) {0};
\node[below, align=center]
at (axis cs:1,3) {0.0\%};

\addplot[area legend, draw=black, fill=mycolor2]
table[row sep=crcr] {%
x	y\\
0.5	3.5\\
1.5	3.5\\
1.5	4.5\\
0.5	4.5\\
}--cycle;

\node[above, align=center, font=\bfseries]
at (axis cs:1,4) {0};
\node[below, align=center]
at (axis cs:1,4) {0.0\%};

\addplot[area legend, draw=black, fill=mycolor2]
table[row sep=crcr] {%
x	y\\
0.5	4.5\\
1.5	4.5\\
1.5	5.5\\
0.5	5.5\\
}--cycle;

\node[above, align=center, font=\bfseries]
at (axis cs:1,5) {0};
\node[below, align=center]
at (axis cs:1,5) {0.0\%};

\addplot[area legend, draw=black, fill=mycolor2]
table[row sep=crcr] {%
x	y\\
0.5	5.5\\
1.5	5.5\\
1.5	6.5\\
0.5	6.5\\
}--cycle;

\node[above, align=center, font=\bfseries]
at (axis cs:1,6) {0};
\node[below, align=center]
at (axis cs:1,6) {0.0\%};

\addplot[area legend, draw=black, fill=mycolor2]
table[row sep=crcr] {%
x	y\\
0.5	6.5\\
1.5	6.5\\
1.5	7.5\\
0.5	7.5\\
}--cycle;

\node[above, align=center, font=\bfseries]
at (axis cs:1,7) {0};
\node[below, align=center]
at (axis cs:1,7) {0.0\%};

\addplot[area legend, draw=black, fill=gray]
table[row sep=crcr] {%
x	y\\
0.5	7.5\\
1.5	7.5\\
1.5	8.5\\
0.5	8.5\\
}--cycle;

\node[above, align=center, font=\color{black!60!green}]
at (axis cs:1,8) {100.0\%};
\node[below, align=center, font=\color{black!60!red}]
at (axis cs:1,8) {0.0\%};

\addplot[area legend, draw=black, fill=mycolor2]
table[row sep=crcr] {%
x	y\\
1.5	0.5\\
2.5	0.5\\
2.5	1.5\\
1.5	1.5\\
}--cycle;

\node[above, align=center, font=\bfseries]
at (axis cs:2,1) {0};
\node[below, align=center]
at (axis cs:2,1) {0.0\%};

\addplot[area legend, draw=black, fill=mycolor1]
table[row sep=crcr] {%
x	y\\
1.5	1.5\\
2.5	1.5\\
2.5	2.5\\
1.5	2.5\\
}--cycle;

\node[above, align=center, font=\bfseries]
at (axis cs:2,2) {22};
\node[below, align=center]
at (axis cs:2,2) {16.3\%};

\addplot[area legend, draw=black, fill=mycolor2]
table[row sep=crcr] {%
x	y\\
1.5	2.5\\
2.5	2.5\\
2.5	3.5\\
1.5	3.5\\
}--cycle;

\node[above, align=center, font=\bfseries]
at (axis cs:2,3) {0};
\node[below, align=center]
at (axis cs:2,3) {0.0\%};

\addplot[area legend, draw=black, fill=mycolor2]
table[row sep=crcr] {%
x	y\\
1.5	3.5\\
2.5	3.5\\
2.5	4.5\\
1.5	4.5\\
}--cycle;

\node[above, align=center, font=\bfseries]
at (axis cs:2,4) {0};
\node[below, align=center]
at (axis cs:2,4) {0.0\%};

\addplot[area legend, draw=black, fill=mycolor2]
table[row sep=crcr] {%
x	y\\
1.5	4.5\\
2.5	4.5\\
2.5	5.5\\
1.5	5.5\\
}--cycle;

\node[above, align=center, font=\bfseries]
at (axis cs:2,5) {0};
\node[below, align=center]
at (axis cs:2,5) {0.0\%};

\addplot[area legend, draw=black, fill=mycolor2]
table[row sep=crcr] {%
x	y\\
1.5	5.5\\
2.5	5.5\\
2.5	6.5\\
1.5	6.5\\
}--cycle;

\node[above, align=center, font=\bfseries]
at (axis cs:2,6) {0};
\node[below, align=center]
at (axis cs:2,6) {0.0\%};

\addplot[area legend, draw=black, fill=mycolor2]
table[row sep=crcr] {%
x	y\\
1.5	6.5\\
2.5	6.5\\
2.5	7.5\\
1.5	7.5\\
}--cycle;

\node[above, align=center, font=\bfseries]
at (axis cs:2,7) {0};
\node[below, align=center]
at (axis cs:2,7) {0.0\%};

\addplot[area legend, draw=black, fill=gray]
table[row sep=crcr] {%
x	y\\
1.5	7.5\\
2.5	7.5\\
2.5	8.5\\
1.5	8.5\\
}--cycle;

\node[above, align=center, font=\color{black!60!green}]
at (axis cs:2,8) {100.0\%};
\node[below, align=center, font=\color{black!60!red}]
at (axis cs:2,8) {0.0\%};

\addplot[area legend, draw=black, fill=mycolor2]
table[row sep=crcr] {%
x	y\\
2.5	0.5\\
3.5	0.5\\
3.5	1.5\\
2.5	1.5\\
}--cycle;

\node[above, align=center, font=\bfseries]
at (axis cs:3,1) {0};
\node[below, align=center]
at (axis cs:3,1) {0.0\%};

\addplot[area legend, draw=black, fill=mycolor2]
table[row sep=crcr] {%
x	y\\
2.5	1.5\\
3.5	1.5\\
3.5	2.5\\
2.5	2.5\\
}--cycle;

\node[above, align=center, font=\bfseries]
at (axis cs:3,2) {0};
\node[below, align=center]
at (axis cs:3,2) {0.0\%};

\addplot[area legend, draw=black, fill=mycolor1]
table[row sep=crcr] {%
x	y\\
2.5	2.5\\
3.5	2.5\\
3.5	3.5\\
2.5	3.5\\
}--cycle;

\node[above, align=center, font=\bfseries]
at (axis cs:3,3) {20};
\node[below, align=center]
at (axis cs:3,3) {15.6\%};

\addplot[area legend, draw=black, fill=mycolor2]
table[row sep=crcr] {%
x	y\\
2.5	3.5\\
3.5	3.5\\
3.5	4.5\\
2.5	4.5\\
}--cycle;

\node[above, align=center, font=\bfseries]
at (axis cs:3,4) {0};
\node[below, align=center]
at (axis cs:3,4) {0.0\%};

\addplot[area legend, draw=black, fill=mycolor2]
table[row sep=crcr] {%
x	y\\
2.5	4.5\\
3.5	4.5\\
3.5	5.5\\
2.5	5.5\\
}--cycle;

\node[above, align=center, font=\bfseries]
at (axis cs:3,5) {0};
\node[below, align=center]
at (axis cs:3,5) {0.0\%};

\addplot[area legend, draw=black, fill=mycolor2]
table[row sep=crcr] {%
x	y\\
2.5	5.5\\
3.5	5.5\\
3.5	6.5\\
2.5	6.5\\
}--cycle;

\node[above, align=center, font=\bfseries]
at (axis cs:3,6) {0};
\node[below, align=center]
at (axis cs:3,6) {0.0\%};

\addplot[area legend, draw=black, fill=mycolor2]
table[row sep=crcr] {%
x	y\\
2.5	6.5\\
3.5	6.5\\
3.5	7.5\\
2.5	7.5\\
}--cycle;

\node[above, align=center, font=\bfseries]
at (axis cs:3,7) {0};
\node[below, align=center]
at (axis cs:3,7) {0.0\%};

\addplot[area legend, draw=black, fill=gray]
table[row sep=crcr] {%
x	y\\
2.5	7.5\\
3.5	7.5\\
3.5	8.5\\
2.5	8.5\\
}--cycle;

\node[above, align=center, font=\color{black!60!green}]
at (axis cs:3,8) {100\%};
\node[below, align=center, font=\color{black!60!red}]
at (axis cs:3,8) {0.0\%};

\addplot[area legend, draw=black, fill=mycolor2]
table[row sep=crcr] {%
x	y\\
3.5	0.5\\
4.5	0.5\\
4.5	1.5\\
3.5	1.5\\
}--cycle;

\node[above, align=center, font=\bfseries]
at (axis cs:4,1) {0};
\node[below, align=center]
at (axis cs:4,1) {0.0\%};

\addplot[area legend, draw=black, fill=mycolor2]
table[row sep=crcr] {%
x	y\\
3.5	1.5\\
4.5	1.5\\
4.5	2.5\\
3.5	2.5\\
}--cycle;

\node[above, align=center, font=\bfseries]
at (axis cs:4,2) {0};
\node[below, align=center]
at (axis cs:4,2) {0.0\%};

\addplot[area legend, draw=black, fill=mycolor2]
table[row sep=crcr] {%
x	y\\
3.5	2.5\\
4.5	2.5\\
4.5	3.5\\
3.5	3.5\\
}--cycle;

\node[above, align=center, font=\bfseries]
at (axis cs:4,3) {0};
\node[below, align=center]
at (axis cs:4,3) {0.0\%};

\addplot[area legend, draw=black, fill=mycolor1]
table[row sep=crcr] {%
x	y\\
3.5	3.5\\
4.5	3.5\\
4.5	4.5\\
3.5	4.5\\
}--cycle;

\node[above, align=center, font=\bfseries]
at (axis cs:4,4) {23};
\node[below, align=center]
at (axis cs:4,4) {18.0\%};

\addplot[area legend, draw=black, fill=mycolor2]
table[row sep=crcr] {%
x	y\\
3.5	4.5\\
4.5	4.5\\
4.5	5.5\\
3.5	5.5\\
}--cycle;

\node[above, align=center, font=\bfseries]
at (axis cs:4,5) {0};
\node[below, align=center]
at (axis cs:4,5) {0.0\%};

\addplot[area legend, draw=black, fill=mycolor2]
table[row sep=crcr] {%
x	y\\
3.5	5.5\\
4.5	5.5\\
4.5	6.5\\
3.5	6.5\\
}--cycle;

\node[above, align=center, font=\bfseries]
at (axis cs:4,6) {0};
\node[below, align=center]
at (axis cs:4,6) {0.0\%};

\addplot[area legend, draw=black, fill=mycolor2]
table[row sep=crcr] {%
x	y\\
3.5	6.5\\
4.5	6.5\\
4.5	7.5\\
3.5	7.5\\
}--cycle;

\node[above, align=center, font=\bfseries]
at (axis cs:4,7) {0};
\node[below, align=center]
at (axis cs:4,7) {0.0\%};

\addplot[area legend, draw=black, fill=gray]
table[row sep=crcr] {%
x	y\\
3.5	7.5\\
4.5	7.5\\
4.5	8.5\\
3.5	8.5\\
}--cycle;

\node[above, align=center, font=\color{black!60!green}]
at (axis cs:4,8) {100\%};
\node[below, align=center, font=\color{black!60!red}]
at (axis cs:4,8) {0.0\%};

\addplot[area legend, draw=black, fill=mycolor2]
table[row sep=crcr] {%
x	y\\
4.5	0.5\\
5.5	0.5\\
5.5	1.5\\
4.5	1.5\\
}--cycle;

\node[above, align=center, font=\bfseries]
at (axis cs:5,1) {0};
\node[below, align=center]
at (axis cs:5,1) {0.0\%};

\addplot[area legend, draw=black, fill=mycolor2]
table[row sep=crcr] {%
x	y\\
4.5	1.5\\
5.5	1.5\\
5.5	2.5\\
4.5	2.5\\
}--cycle;

\node[above, align=center, font=\bfseries]
at (axis cs:5,2) {0};
\node[below, align=center]
at (axis cs:5,2) {0.0\%};

\addplot[area legend, draw=black, fill=mycolor2]
table[row sep=crcr] {%
x	y\\
4.5	2.5\\
5.5	2.5\\
5.5	3.5\\
4.5	3.5\\
}--cycle;

\node[above, align=center, font=\bfseries]
at (axis cs:5,3) {0};
\node[below, align=center]
at (axis cs:5,3) {0.0\%};

\addplot[area legend, draw=black, fill=mycolor2]
table[row sep=crcr] {%
x	y\\
4.5	3.5\\
5.5	3.5\\
5.5	4.5\\
4.5	4.5\\
}--cycle;

\node[above, align=center, font=\bfseries]
at (axis cs:5,4) {0};
\node[below, align=center]
at (axis cs:5,4) {0.0\%};

\addplot[area legend, draw=black, fill=mycolor1]
table[row sep=crcr] {%
x	y\\
4.5	4.5\\
5.5	4.5\\
5.5	5.5\\
4.5	5.5\\
}--cycle;

\node[above, align=center, font=\bfseries]
at (axis cs:5,5) {20};
\node[below, align=center]
at (axis cs:5,5) {15.6\%};

\addplot[area legend, draw=black, fill=mycolor2]
table[row sep=crcr] {%
x	y\\
4.5	5.5\\
5.5	5.5\\
5.5	6.5\\
4.5	6.5\\
}--cycle;

\node[above, align=center, font=\bfseries]
at (axis cs:5,6) {0};
\node[below, align=center]
at (axis cs:5,6) {0.0\%};

\addplot[area legend, draw=black, fill=mycolor2]
table[row sep=crcr] {%
x	y\\
4.5	6.5\\
5.5	6.5\\
5.5	7.5\\
4.5	7.5\\
}--cycle;

\node[above, align=center, font=\bfseries]
at (axis cs:5,7) {1};
\node[below, align=center]
at (axis cs:5,7) {0.8\%};

\addplot[area legend, draw=black, fill=gray]
table[row sep=crcr] {%
x	y\\
4.5	7.5\\
5.5	7.5\\
5.5	8.5\\
4.5	8.5\\
}--cycle;

\node[above, align=center, font=\color{black!60!green}]
at (axis cs:5,8) {95.2\%};
\node[below, align=center, font=\color{black!60!red}]
at (axis cs:5,8) {4.8\%};

\addplot[area legend, draw=black, fill=mycolor2]
table[row sep=crcr] {%
x	y\\
5.5	0.5\\
6.5	0.5\\
6.5	1.5\\
5.5	1.5\\
}--cycle;

\node[above, align=center, font=\bfseries]
at (axis cs:6,1) {0};
\node[below, align=center]
at (axis cs:6,1) {0.0\%};

\addplot[area legend, draw=black, fill=mycolor2]
table[row sep=crcr] {%
x	y\\
5.5	1.5\\
6.5	1.5\\
6.5	2.5\\
5.5	2.5\\
}--cycle;

\node[above, align=center, font=\bfseries]
at (axis cs:6,2) {0};
\node[below, align=center]
at (axis cs:6,2) {0.0\%};

\addplot[area legend, draw=black, fill=mycolor2]
table[row sep=crcr] {%
x	y\\
5.5	2.5\\
6.5	2.5\\
6.5	3.5\\
5.5	3.5\\
}--cycle;

\node[above, align=center, font=\bfseries]
at (axis cs:6,3) {0};
\node[below, align=center]
at (axis cs:6,3) {0.0\%};

\addplot[area legend, draw=black, fill=mycolor2]
table[row sep=crcr] {%
x	y\\
5.5	3.5\\
6.5	3.5\\
6.5	4.5\\
5.5	4.5\\
}--cycle;

\node[above, align=center, font=\bfseries]
at (axis cs:6,4) {0};
\node[below, align=center]
at (axis cs:6,4) {0.0\%};

\addplot[area legend, draw=black, fill=mycolor2]
table[row sep=crcr] {%
x	y\\
5.5	4.5\\
6.5	4.5\\
6.5	5.5\\
5.5	5.5\\
}--cycle;

\node[above, align=center, font=\bfseries]
at (axis cs:6,5) {2};
\node[below, align=center]
at (axis cs:6,5) {1.6\%};

\addplot[area legend, draw=black, fill=mycolor1]
table[row sep=crcr] {%
x	y\\
5.5	5.5\\
6.5	5.5\\
6.5	6.5\\
5.5	6.5\\
}--cycle;

\node[above, align=center, font=\bfseries]
at (axis cs:6,6) {14};
\node[below, align=center]
at (axis cs:6,6) {10.9\%};

\addplot[area legend, draw=black, fill=mycolor2]
table[row sep=crcr] {%
x	y\\
5.5	6.5\\
6.5	6.5\\
6.5	7.5\\
5.5	7.5\\
}--cycle;

\node[above, align=center, font=\bfseries]
at (axis cs:6,7) {6};
\node[below, align=center]
at (axis cs:6,7) {4.7\%};

\addplot[area legend, draw=black, fill=gray]
table[row sep=crcr] {%
x	y\\
5.5	7.5\\
6.5	7.5\\
6.5	8.5\\
5.5	8.5\\
}--cycle;

\node[above, align=center, font=\color{black!60!green}]
at (axis cs:6,8) {63.6\%};
\node[below, align=center, font=\color{black!60!red}]
at (axis cs:6,8) {36.4\%};

\addplot[area legend, draw=black, fill=mycolor2]
table[row sep=crcr] {%
x	y\\
6.5	0.5\\
7.5	0.5\\
7.5	1.5\\
6.5	1.5\\
}--cycle;

\node[above, align=center, font=\bfseries]
at (axis cs:7,1) {1};
\node[below, align=center]
at (axis cs:7,1) {0.7\%};

\addplot[area legend, draw=black, fill=mycolor2]
table[row sep=crcr] {%
x	y\\
6.5	1.5\\
7.5	1.5\\
7.5	2.5\\
6.5	2.5\\
}--cycle;

\node[above, align=center, font=\bfseries]
at (axis cs:7,2) {0};
\node[below, align=center]
at (axis cs:7,2) {0.0\%};

\addplot[area legend, draw=black, fill=mycolor2, forget plot]
table[row sep=crcr] {%
x	y\\
6.5	2.5\\
7.5	2.5\\
7.5	3.5\\
6.5	3.5\\
}--cycle;
\node[above, align=center, font=\bfseries]
at (axis cs:7,3) {1};
\node[below, align=center]
at (axis cs:7,3) {0.7\%};

\addplot[area legend, draw=black, fill=mycolor2, forget plot]
table[row sep=crcr] {%
x	y\\
6.5	3.5\\
7.5	3.5\\
7.5	4.5\\
6.5	4.5\\
}--cycle;
\node[above, align=center, font=\bfseries]
at (axis cs:7,4) {0};
\node[below, align=center]
at (axis cs:7,4) {0.0\%};

\addplot[area legend, draw=black, fill=mycolor2, forget plot]
table[row sep=crcr] {%
x	y\\
6.5	4.5\\
7.5	4.5\\
7.5	5.5\\
6.5	5.5\\
}--cycle;
\node[above, align=center, font=\bfseries]
at (axis cs:7,5) {3};
\node[below, align=center]
at (axis cs:7,5) {2.2\%};

\addplot[area legend, draw=black, fill=mycolor2, forget plot]
table[row sep=crcr] {%
x	y\\
6.5	5.5\\
7.5	5.5\\
7.5	6.5\\
6.5	6.5\\
}--cycle;
\node[above, align=center, font=\bfseries]
at (axis cs:7,6) {2};
\node[below, align=center]
at (axis cs:7,6) {1.5\%};

\addplot[area legend, draw=black, fill=mycolor1, forget plot]
table[row sep=crcr] {%
x	y\\
6.5	6.5\\
7.5	6.5\\
7.5	7.5\\
6.5	7.5\\
}--cycle;
\node[above, align=center, font=\bfseries]
at (axis cs:7,7) {0};
\node[below, align=center]
at (axis cs:7,7) {0.0\%};

\addplot[area legend, draw=black, fill=gray, forget plot]
table[row sep=crcr] {%
x	y\\
6.5	7.5\\
7.5	7.5\\
7.5	8.5\\
6.5	8.5\\
}--cycle;
\node[above, align=center, font=\color{black!60!green}]
at (axis cs:7,8) {0.0\%};
\node[below, align=center, font=\color{black!60!red}]
at (axis cs:7,8) {100.0\%};

\addplot[area legend, draw=black, fill=gray, forget plot]
table[row sep=crcr] {%
x	y\\
7.5	0.5\\
8.5	0.5\\
8.5	1.5\\
7.5	1.5\\
}--cycle;
\node[above, align=center, font=\color{black!60!green}]
at (axis cs:8,1) {95.2\%};
\node[below, align=center, font=\color{black!60!red}]
at (axis cs:8,1) {4.8\%};

\addplot[area legend, draw=black, fill=gray, forget plot]
table[row sep=crcr] {%
x	y\\
7.5	1.5\\
8.5	1.5\\
8.5	2.5\\
7.5	2.5\\
}--cycle;
\node[above, align=center, font=\color{black!60!green}]
at (axis cs:8,2) {100\%};
\node[below, align=center, font=\color{black!60!red}]
at (axis cs:8,2) {0.0\%};

\addplot[area legend, draw=black, fill=gray, forget plot]
table[row sep=crcr] {%
x	y\\
7.5	2.5\\
8.5	2.5\\
8.5	3.5\\
7.5	3.5\\
}--cycle;
\node[above, align=center, font=\color{black!60!green}]
at (axis cs:8,3) {95.2\%};
\node[below, align=center, font=\color{black!60!red}]
at (axis cs:8,3) {4.8\%};

\addplot[area legend, draw=black, fill=gray, forget plot]
table[row sep=crcr] {%
x	y\\
7.5	3.5\\
8.5	3.5\\
8.5	4.5\\
7.5	4.5\\
}--cycle;
\node[above, align=center, font=\color{black!60!green}]
at (axis cs:8,4) {100\%};
\node[below, align=center, font=\color{black!60!red}]
at (axis cs:8,4) {0.0\%};

\addplot[area legend, draw=black, fill=gray, forget plot]
table[row sep=crcr] {%
x	y\\
7.5	4.5\\
8.5	4.5\\
8.5	5.5\\
7.5	5.5\\
}--cycle;
\node[above, align=center, font=\color{black!60!green}]
at (axis cs:8,5) {80.0\%};
\node[below, align=center, font=\color{black!60!red}]
at (axis cs:8,5) {20.0\%};

\addplot[area legend, draw=black, fill=gray, forget plot]
table[row sep=crcr] {%
x	y\\
7.5	5.5\\
8.5	5.5\\
8.5	6.5\\
7.5	6.5\\
}--cycle;
\node[above, align=center, font=\color{black!60!green}]
at (axis cs:8,6) {87.5\%};
\node[below, align=center, font=\color{black!60!red}]
at (axis cs:8,6) {12.5\%};

\addplot[area legend, draw=black, fill=gray, forget plot]
table[row sep=crcr] {%
x	y\\
7.5	6.5\\
8.5	6.5\\
8.5	7.5\\
7.5	7.5\\
}--cycle;
\node[above, align=center, font=\color{black!60!green}]
at (axis cs:8,7) {0.0\%};
\node[below, align=center, font=\color{black!60!red}]
at (axis cs:8,7) {100\%};

\addplot[area legend, draw=black, fill=mycolor3, forget plot]
table[row sep=crcr] {%
x	y\\
7.5	7.5\\
8.5	7.5\\
8.5	8.5\\
7.5	8.5\\
}--cycle;
\node[above, align=center, font=\bfseries\color{black!60!green}]
at (axis cs:8,8) {88.1\%};
\node[below, align=center, font=\bfseries\color{black!60!red}]
at (axis cs:8,8) {11.9\%};
\addplot [color=darkgray, line width=2.0pt, forget plot]
  table[row sep=crcr]{%
7.5	0.5\\
7.5	8.5\\
};
\addplot [color=darkgray, line width=2.0pt, forget plot]
  table[row sep=crcr]{%
0.5	7.5\\
8.5	7.5\\
};
\end{axis}
\end{tikzpicture}%

%% file: confusion_matrix_online_60.tex
%
%
\definecolor{mycolor1}{rgb}{0.47059,0.90196,0.70588}%
\definecolor{mycolor2}{rgb}{0.90196,0.54902,0.54902}%
\definecolor{mycolor3}{rgb}{0.47059,0.58824,0.90196}%
\begin{tikzpicture}

\begin{axis}[%
width=\figurewidth,
height=\figureheight,
at={(4.795in,0.693in)},
scale only axis,
xmin=0.5,
xmax=8.5,
xtick={1,2,3,4,5,6,7,8,9},
xticklabels={{$\text{G}_\text{1}$},{$\text{G}_\text{2}$},{$\text{G}_\text{3}$},{$\text{G}_\text{4}$},{$\text{G}_\text{5}$},{$\text{G}_\text{6}$},{N.G.},{},{1}},
xlabel style={font=\color{white!15!black}},
xlabel={Target Class},
y dir=reverse,
ymin=0.5,
ymax=8.5,
ytick={1,2,3,4,5,6,7,8,9},
yticklabels={{$\text{G}_\text{1}$},{$\text{G}_\text{2}$},{$\text{G}_\text{3}$},{$\text{G}_\text{4}$},{$\text{G}_\text{5}$},{$\text{G}_\text{6}$},{N.G.},{},{1}},
ylabel style={font=\color{white!15!black}},
ylabel={Output Class},
axis background/.style={fill=white},
title style={font=\bfseries},
title={ Confusion Matrix},
legend style={legend cell align=left, align=left, draw=white!15!black}
]

\addplot[area legend, draw=black, fill=mycolor1]
table[row sep=crcr] {%
x	y\\
0.5	0.5\\
1.5	0.5\\
1.5	1.5\\
0.5	1.5\\
}--cycle;

\node[above, align=center, font=\bfseries]
at (axis cs:1,1) {20};
\node[below, align=center]
at (axis cs:1,1) {14.5\%};

\addplot[area legend, draw=black, fill=mycolor2]
table[row sep=crcr] {%
x	y\\
0.5	1.5\\
1.5	1.5\\
1.5	2.5\\
0.5	2.5\\
}--cycle;

\node[above, align=center, font=\bfseries]
at (axis cs:1,2) {0};
\node[below, align=center]
at (axis cs:1,2) {0.0\%};

\addplot[area legend, draw=black, fill=mycolor2]
table[row sep=crcr] {%
x	y\\
0.5	2.5\\
1.5	2.5\\
1.5	3.5\\
0.5	3.5\\
}--cycle;

\node[above, align=center, font=\bfseries]
at (axis cs:1,3) {0};
\node[below, align=center]
at (axis cs:1,3) {0.0\%};

\addplot[area legend, draw=black, fill=mycolor2]
table[row sep=crcr] {%
x	y\\
0.5	3.5\\
1.5	3.5\\
1.5	4.5\\
0.5	4.5\\
}--cycle;

\node[above, align=center, font=\bfseries]
at (axis cs:1,4) {0};
\node[below, align=center]
at (axis cs:1,4) {0.0\%};

\addplot[area legend, draw=black, fill=mycolor2]
table[row sep=crcr] {%
x	y\\
0.5	4.5\\
1.5	4.5\\
1.5	5.5\\
0.5	5.5\\
}--cycle;

\node[above, align=center, font=\bfseries]
at (axis cs:1,5) {0};
\node[below, align=center]
at (axis cs:1,5) {0.0\%};

\addplot[area legend, draw=black, fill=mycolor2]
table[row sep=crcr] {%
x	y\\
0.5	5.5\\
1.5	5.5\\
1.5	6.5\\
0.5	6.5\\
}--cycle;

\node[above, align=center, font=\bfseries]
at (axis cs:1,6) {0};
\node[below, align=center]
at (axis cs:1,6) {0.0\%};

\addplot[area legend, draw=black, fill=mycolor2]
table[row sep=crcr] {%
x	y\\
0.5	6.5\\
1.5	6.5\\
1.5	7.5\\
0.5	7.5\\
}--cycle;

\node[above, align=center, font=\bfseries]
at (axis cs:1,7) {0};
\node[below, align=center]
at (axis cs:1,7) {0.0\%};

\addplot[area legend, draw=black, fill=gray]
table[row sep=crcr] {%
x	y\\
0.5	7.5\\
1.5	7.5\\
1.5	8.5\\
0.5	8.5\\
}--cycle;

\node[above, align=center, font=\color{black!60!green}]
at (axis cs:1,8) {100.0\%};
\node[below, align=center, font=\color{black!60!red}]
at (axis cs:1,8) {0.0\%};

\addplot[area legend, draw=black, fill=mycolor2]
table[row sep=crcr] {%
x	y\\
1.5	0.5\\
2.5	0.5\\
2.5	1.5\\
1.5	1.5\\
}--cycle;

\node[above, align=center, font=\bfseries]
at (axis cs:2,1) {0};
\node[below, align=center]
at (axis cs:2,1) {0.0\%};

\addplot[area legend, draw=black, fill=mycolor1]
table[row sep=crcr] {%
x	y\\
1.5	1.5\\
2.5	1.5\\
2.5	2.5\\
1.5	2.5\\
}--cycle;

\node[above, align=center, font=\bfseries]
at (axis cs:2,2) {22};
\node[below, align=center]
at (axis cs:2,2) {15.9\%};

\addplot[area legend, draw=black, fill=mycolor2]
table[row sep=crcr] {%
x	y\\
1.5	2.5\\
2.5	2.5\\
2.5	3.5\\
1.5	3.5\\
}--cycle;

\node[above, align=center, font=\bfseries]
at (axis cs:2,3) {0};
\node[below, align=center]
at (axis cs:2,3) {0.0\%};

\addplot[area legend, draw=black, fill=mycolor2]
table[row sep=crcr] {%
x	y\\
1.5	3.5\\
2.5	3.5\\
2.5	4.5\\
1.5	4.5\\
}--cycle;

\node[above, align=center, font=\bfseries]
at (axis cs:2,4) {0};
\node[below, align=center]
at (axis cs:2,4) {0.0\%};

\addplot[area legend, draw=black, fill=mycolor2]
table[row sep=crcr] {%
x	y\\
1.5	4.5\\
2.5	4.5\\
2.5	5.5\\
1.5	5.5\\
}--cycle;

\node[above, align=center, font=\bfseries]
at (axis cs:2,5) {0};
\node[below, align=center]
at (axis cs:2,5) {0.0\%};

\addplot[area legend, draw=black, fill=mycolor2]
table[row sep=crcr] {%
x	y\\
1.5	5.5\\
2.5	5.5\\
2.5	6.5\\
1.5	6.5\\
}--cycle;

\node[above, align=center, font=\bfseries]
at (axis cs:2,6) {0};
\node[below, align=center]
at (axis cs:2,6) {0.0\%};

\addplot[area legend, draw=black, fill=mycolor2]
table[row sep=crcr] {%
x	y\\
1.5	6.5\\
2.5	6.5\\
2.5	7.5\\
1.5	7.5\\
}--cycle;

\node[above, align=center, font=\bfseries]
at (axis cs:2,7) {0};
\node[below, align=center]
at (axis cs:2,7) {0.0\%};

\addplot[area legend, draw=black, fill=gray]
table[row sep=crcr] {%
x	y\\
1.5	7.5\\
2.5	7.5\\
2.5	8.5\\
1.5	8.5\\
}--cycle;

\node[above, align=center, font=\color{black!60!green}]
at (axis cs:2,8) {100\%};
\node[below, align=center, font=\color{black!60!red}]
at (axis cs:2,8) {0.0\%};

\addplot[area legend, draw=black, fill=mycolor2]
table[row sep=crcr] {%
x	y\\
2.5	0.5\\
3.5	0.5\\
3.5	1.5\\
2.5	1.5\\
}--cycle;

\node[above, align=center, font=\bfseries]
at (axis cs:3,1) {0};
\node[below, align=center]
at (axis cs:3,1) {0.0\%};

\addplot[area legend, draw=black, fill=mycolor2]
table[row sep=crcr] {%
x	y\\
2.5	1.5\\
3.5	1.5\\
3.5	2.5\\
2.5	2.5\\
}--cycle;

\node[above, align=center, font=\bfseries]
at (axis cs:3,2) {0};
\node[below, align=center]
at (axis cs:3,2) {0.0\%};

\addplot[area legend, draw=black, fill=mycolor1]
table[row sep=crcr] {%
x	y\\
2.5	2.5\\
3.5	2.5\\
3.5	3.5\\
2.5	3.5\\
}--cycle;

\node[above, align=center, font=\bfseries]
at (axis cs:3,3) {20};
\node[below, align=center]
at (axis cs:3,3) {14.5\%};

\addplot[area legend, draw=black, fill=mycolor2]
table[row sep=crcr] {%
x	y\\
2.5	3.5\\
3.5	3.5\\
3.5	4.5\\
2.5	4.5\\
}--cycle;

\node[above, align=center, font=\bfseries]
at (axis cs:3,4) {0};
\node[below, align=center]
at (axis cs:3,4) {0.0\%};

\addplot[area legend, draw=black, fill=mycolor2]
table[row sep=crcr] {%
x	y\\
2.5	4.5\\
3.5	4.5\\
3.5	5.5\\
2.5	5.5\\
}--cycle;

\node[above, align=center, font=\bfseries]
at (axis cs:3,5) {0};
\node[below, align=center]
at (axis cs:3,5) {0.0\%};

\addplot[area legend, draw=black, fill=mycolor2]
table[row sep=crcr] {%
x	y\\
2.5	5.5\\
3.5	5.5\\
3.5	6.5\\
2.5	6.5\\
}--cycle;

\node[above, align=center, font=\bfseries]
at (axis cs:3,6) {0};
\node[below, align=center]
at (axis cs:3,6) {0.0\%};

\addplot[area legend, draw=black, fill=mycolor2]
table[row sep=crcr] {%
x	y\\
2.5	6.5\\
3.5	6.5\\
3.5	7.5\\
2.5	7.5\\
}--cycle;

\node[above, align=center, font=\bfseries]
at (axis cs:3,7) {0};
\node[below, align=center]
at (axis cs:3,7) {0.0\%};

\addplot[area legend, draw=black, fill=gray]
table[row sep=crcr] {%
x	y\\
2.5	7.5\\
3.5	7.5\\
3.5	8.5\\
2.5	8.5\\
}--cycle;

\node[above, align=center, font=\color{black!60!green}]
at (axis cs:3,8) {100.0\%};
\node[below, align=center, font=\color{black!60!red}]
at (axis cs:3,8) {0.0\%};

\addplot[area legend, draw=black, fill=mycolor2]
table[row sep=crcr] {%
x	y\\
3.5	0.5\\
4.5	0.5\\
4.5	1.5\\
3.5	1.5\\
}--cycle;

\node[above, align=center, font=\bfseries]
at (axis cs:4,1) {0};
\node[below, align=center]
at (axis cs:4,1) {0.0\%};

\addplot[area legend, draw=black, fill=mycolor2]
table[row sep=crcr] {%
x	y\\
3.5	1.5\\
4.5	1.5\\
4.5	2.5\\
3.5	2.5\\
}--cycle;

\node[above, align=center, font=\bfseries]
at (axis cs:4,2) {0};
\node[below, align=center]
at (axis cs:4,2) {0.0\%};

\addplot[area legend, draw=black, fill=mycolor2]
table[row sep=crcr] {%
x	y\\
3.5	2.5\\
4.5	2.5\\
4.5	3.5\\
3.5	3.5\\
}--cycle;

\node[above, align=center, font=\bfseries]
at (axis cs:4,3) {0};
\node[below, align=center]
at (axis cs:4,3) {0.0\%};

\addplot[area legend, draw=black, fill=mycolor1]
table[row sep=crcr] {%
x	y\\
3.5	3.5\\
4.5	3.5\\
4.5	4.5\\
3.5	4.5\\
}--cycle;

\node[above, align=center, font=\bfseries]
at (axis cs:4,4) {23};
\node[below, align=center]
at (axis cs:4,4) {15.2\%};

\addplot[area legend, draw=black, fill=mycolor2]
table[row sep=crcr] {%
x	y\\
3.5	4.5\\
4.5	4.5\\
4.5	5.5\\
3.5	5.5\\
}--cycle;

\node[above, align=center, font=\bfseries]
at (axis cs:4,5) {0};
\node[below, align=center]
at (axis cs:4,5) {0.0\%};

\addplot[area legend, draw=black, fill=mycolor2]
table[row sep=crcr] {%
x	y\\
3.5	5.5\\
4.5	5.5\\
4.5	6.5\\
3.5	6.5\\
}--cycle;

\node[above, align=center, font=\bfseries]
at (axis cs:4,6) {0};
\node[below, align=center]
at (axis cs:4,6) {0.0\%};

\addplot[area legend, draw=black, fill=mycolor2]
table[row sep=crcr] {%
x	y\\
3.5	6.5\\
4.5	6.5\\
4.5	7.5\\
3.5	7.5\\
}--cycle;

\node[above, align=center, font=\bfseries]
at (axis cs:4,7) {0};
\node[below, align=center]
at (axis cs:4,7) {0.0\%};

\addplot[area legend, draw=black, fill=gray]
table[row sep=crcr] {%
x	y\\
3.5	7.5\\
4.5	7.5\\
4.5	8.5\\
3.5	8.5\\
}--cycle;

\node[above, align=center, font=\color{black!60!green}]
at (axis cs:4,8) {100.0\%};
\node[below, align=center, font=\color{black!60!red}]
at (axis cs:4,8) {0.0\%};

\addplot[area legend, draw=black, fill=mycolor2]
table[row sep=crcr] {%
x	y\\
4.5	0.5\\
5.5	0.5\\
5.5	1.5\\
4.5	1.5\\
}--cycle;

\node[above, align=center, font=\bfseries]
at (axis cs:5,1) {0};
\node[below, align=center]
at (axis cs:5,1) {0.0\%};

\addplot[area legend, draw=black, fill=mycolor2]
table[row sep=crcr] {%
x	y\\
4.5	1.5\\
5.5	1.5\\
5.5	2.5\\
4.5	2.5\\
}--cycle;

\node[above, align=center, font=\bfseries]
at (axis cs:5,2) {0};
\node[below, align=center]
at (axis cs:5,2) {0.0\%};

\addplot[area legend, draw=black, fill=mycolor2]
table[row sep=crcr] {%
x	y\\
4.5	2.5\\
5.5	2.5\\
5.5	3.5\\
4.5	3.5\\
}--cycle;

\node[above, align=center, font=\bfseries]
at (axis cs:5,3) {0};
\node[below, align=center]
at (axis cs:5,3) {0.0\%};

\addplot[area legend, draw=black, fill=mycolor2]
table[row sep=crcr] {%
x	y\\
4.5	3.5\\
5.5	3.5\\
5.5	4.5\\
4.5	4.5\\
}--cycle;

\node[above, align=center, font=\bfseries]
at (axis cs:5,4) {0};
\node[below, align=center]
at (axis cs:5,4) {0.0\%};

\addplot[area legend, draw=black, fill=mycolor1]
table[row sep=crcr] {%
x	y\\
4.5	4.5\\
5.5	4.5\\
5.5	5.5\\
4.5	5.5\\
}--cycle;

\node[above, align=center, font=\bfseries]
at (axis cs:5,5) {21};
\node[below, align=center]
at (axis cs:5,5) {15.2\%};

\addplot[area legend, draw=black, fill=mycolor2]
table[row sep=crcr] {%
x	y\\
4.5	5.5\\
5.5	5.5\\
5.5	6.5\\
4.5	6.5\\
}--cycle;

\node[above, align=center, font=\bfseries]
at (axis cs:5,6) {0};
\node[below, align=center]
at (axis cs:5,6) {0.0\%};

\addplot[area legend, draw=black, fill=mycolor2]
table[row sep=crcr] {%
x	y\\
4.5	6.5\\
5.5	6.5\\
5.5	7.5\\
4.5	7.5\\
}--cycle;

\node[above, align=center, font=\bfseries]
at (axis cs:5,7) {0};
\node[below, align=center]
at (axis cs:5,7) {0.0\%};

\addplot[area legend, draw=black, fill=gray]
table[row sep=crcr] {%
x	y\\
4.5	7.5\\
5.5	7.5\\
5.5	8.5\\
4.5	8.5\\
}--cycle;

\node[above, align=center, font=\color{black!60!green}]
at (axis cs:5,8) {100\%};
\node[below, align=center, font=\color{black!60!red}]
at (axis cs:5,8) {0.0\%};

\addplot[area legend, draw=black, fill=mycolor2]
table[row sep=crcr] {%
x	y\\
5.5	0.5\\
6.5	0.5\\
6.5	1.5\\
5.5	1.5\\
}--cycle;

\node[above, align=center, font=\bfseries]
at (axis cs:6,1) {0};
\node[below, align=center]
at (axis cs:6,1) {0.0\%};

\addplot[area legend, draw=black, fill=mycolor2]
table[row sep=crcr] {%
x	y\\
5.5	1.5\\
6.5	1.5\\
6.5	2.5\\
5.5	2.5\\
}--cycle;

\node[above, align=center, font=\bfseries]
at (axis cs:6,2) {0};
\node[below, align=center]
at (axis cs:6,2) {0.0\%};

\addplot[area legend, draw=black, fill=mycolor2]
table[row sep=crcr] {%
x	y\\
5.5	2.5\\
6.5	2.5\\
6.5	3.5\\
5.5	3.5\\
}--cycle;

\node[above, align=center, font=\bfseries]
at (axis cs:6,3) {0};
\node[below, align=center]
at (axis cs:6,3) {0.0\%};

\addplot[area legend, draw=black, fill=mycolor2]
table[row sep=crcr] {%
x	y\\
5.5	3.5\\
6.5	3.5\\
6.5	4.5\\
5.5	4.5\\
}--cycle;

\node[above, align=center, font=\bfseries]
at (axis cs:6,4) {0};
\node[below, align=center]
at (axis cs:6,4) {0.0\%};

\addplot[area legend, draw=black, fill=mycolor2]
table[row sep=crcr] {%
x	y\\
5.5	4.5\\
6.5	4.5\\
6.5	5.5\\
5.5	5.5\\
}--cycle;

\node[above, align=center, font=\bfseries]
at (axis cs:6,5) {1};
\node[below, align=center]
at (axis cs:6,5) {0.7\%};

\addplot[area legend, draw=black, fill=mycolor1]
table[row sep=crcr] {%
x	y\\
5.5	5.5\\
6.5	5.5\\
6.5	6.5\\
5.5	6.5\\
}--cycle;

\node[above, align=center, font=\bfseries]
at (axis cs:6,6) {19};
\node[below, align=center]
at (axis cs:6,6) {13.8\%};

\addplot[area legend, draw=black, fill=mycolor2]
table[row sep=crcr] {%
x	y\\
5.5	6.5\\
6.5	6.5\\
6.5	7.5\\
5.5	7.5\\
}--cycle;

\node[above, align=center, font=\bfseries]
at (axis cs:6,7) {2};
\node[below, align=center]
at (axis cs:6,7) {1.4\%};

\addplot[area legend, draw=black, fill=gray]
table[row sep=crcr] {%
x	y\\
5.5	7.5\\
6.5	7.5\\
6.5	8.5\\
5.5	8.5\\
}--cycle;

\node[above, align=center, font=\color{black!60!green}]
at (axis cs:6,8) {86.4\%};
\node[below, align=center, font=\color{black!60!red}]
at (axis cs:6,8) {13.6\%};

\addplot[area legend, draw=black, fill=mycolor2]
table[row sep=crcr] {%
x	y\\
6.5	0.5\\
7.5	0.5\\
7.5	1.5\\
6.5	1.5\\
}--cycle;

\node[above, align=center, font=\bfseries]
at (axis cs:7,1) {0};
\node[below, align=center]
at (axis cs:7,1) {0.0\%};

\addplot[area legend, draw=black, fill=mycolor2]
table[row sep=crcr] {%
x	y\\
6.5	1.5\\
7.5	1.5\\
7.5	2.5\\
6.5	2.5\\
}--cycle;

\node[above, align=center, font=\bfseries]
at (axis cs:7,2) {0};
\node[below, align=center]
at (axis cs:7,2) {0.0\%};

\addplot[area legend, draw=black, fill=mycolor2, forget plot]
table[row sep=crcr] {%
x	y\\
6.5	2.5\\
7.5	2.5\\
7.5	3.5\\
6.5	3.5\\
}--cycle;
\node[above, align=center, font=\bfseries]
at (axis cs:7,3) {0};
\node[below, align=center]
at (axis cs:7,3) {0.0\%};

\addplot[area legend, draw=black, fill=mycolor2, forget plot]
table[row sep=crcr] {%
x	y\\
6.5	3.5\\
7.5	3.5\\
7.5	4.5\\
6.5	4.5\\
}--cycle;
\node[above, align=center, font=\bfseries]
at (axis cs:7,4) {0};
\node[below, align=center]
at (axis cs:7,4) {0.0\%};

\addplot[area legend, draw=black, fill=mycolor2, forget plot]
table[row sep=crcr] {%
x	y\\
6.5	4.5\\
7.5	4.5\\
7.5	5.5\\
6.5	5.5\\
}--cycle;
\node[above, align=center, font=\bfseries]
at (axis cs:7,5) {1};
\node[below, align=center]
at (axis cs:7,5) {0.7\%};

\addplot[area legend, draw=black, fill=mycolor2, forget plot]
table[row sep=crcr] {%
x	y\\
6.5	5.5\\
7.5	5.5\\
7.5	6.5\\
6.5	6.5\\
}--cycle;
\node[above, align=center, font=\bfseries]
at (axis cs:7,6) {9};
\node[below, align=center]
at (axis cs:7,6) {6.5\%};

\addplot[area legend, draw=black, fill=mycolor1, forget plot]
table[row sep=crcr] {%
x	y\\
6.5	6.5\\
7.5	6.5\\
7.5	7.5\\
6.5	7.5\\
}--cycle;
\node[above, align=center, font=\bfseries]
at (axis cs:7,7) {0};
\node[below, align=center]
at (axis cs:7,7) {0.0\%};

\addplot[area legend, draw=black, fill=gray, forget plot]
table[row sep=crcr] {%
x	y\\
6.5	7.5\\
7.5	7.5\\
7.5	8.5\\
6.5	8.5\\
}--cycle;
\node[above, align=center, font=\color{black!60!green}]
at (axis cs:7,8) {0.0\%};
\node[below, align=center, font=\color{black!60!red}]
at (axis cs:7,8) {100\%};

\addplot[area legend, draw=black, fill=gray, forget plot]
table[row sep=crcr] {%
x	y\\
7.5	0.5\\
8.5	0.5\\
8.5	1.5\\
7.5	1.5\\
}--cycle;
\node[above, align=center, font=\color{black!60!green}]
at (axis cs:8,1) {100\%};
\node[below, align=center, font=\color{black!60!red}]
at (axis cs:8,1) {0.0\%};

\addplot[area legend, draw=black, fill=gray, forget plot]
table[row sep=crcr] {%
x	y\\
7.5	1.5\\
8.5	1.5\\
8.5	2.5\\
7.5	2.5\\
}--cycle;
\node[above, align=center, font=\color{black!60!green}]
at (axis cs:8,2) {100\%};
\node[below, align=center, font=\color{black!60!red}]
at (axis cs:8,2) {0.0\%};

\addplot[area legend, draw=black, fill=gray, forget plot]
table[row sep=crcr] {%
x	y\\
7.5	2.5\\
8.5	2.5\\
8.5	3.5\\
7.5	3.5\\
}--cycle;
\node[above, align=center, font=\color{black!60!green}]
at (axis cs:8,3) {100\%};
\node[below, align=center, font=\color{black!60!red}]
at (axis cs:8,3) {0.0\%};

\addplot[area legend, draw=black, fill=gray, forget plot]
table[row sep=crcr] {%
x	y\\
7.5	3.5\\
8.5	3.5\\
8.5	4.5\\
7.5	4.5\\
}--cycle;
\node[above, align=center, font=\color{black!60!green}]
at (axis cs:8,4) {100\%};
\node[below, align=center, font=\color{black!60!red}]
at (axis cs:8,4) {0.0\%};

\addplot[area legend, draw=black, fill=gray, forget plot]
table[row sep=crcr] {%
x	y\\
7.5	4.5\\
8.5	4.5\\
8.5	5.5\\
7.5	5.5\\
}--cycle;
\node[above, align=center, font=\color{black!60!green}]
at (axis cs:8,5) {91.3\%};
\node[below, align=center, font=\color{black!60!red}]
at (axis cs:8,5) {8.7\%};

\addplot[area legend, draw=black, fill=gray, forget plot]
table[row sep=crcr] {%
x	y\\
7.5	5.5\\
8.5	5.5\\
8.5	6.5\\
7.5	6.5\\
}--cycle;
\node[above, align=center, font=\color{black!60!green}]
at (axis cs:8,6) {67.9\%};
\node[below, align=center, font=\color{black!60!red}]
at (axis cs:8,6) {32.1\%};

\addplot[area legend, draw=black, fill=gray, forget plot]
table[row sep=crcr] {%
x	y\\
7.5	6.5\\
8.5	6.5\\
8.5	7.5\\
7.5	7.5\\
}--cycle;
\node[above, align=center, font=\color{black!60!green}]
at (axis cs:8,7) {0.0\%};
\node[below, align=center, font=\color{black!60!red}]
at (axis cs:8,7) {100\%};

\addplot[area legend, draw=black, fill=mycolor3, forget plot]
table[row sep=crcr] {%
x	y\\
7.5	7.5\\
8.5	7.5\\
8.5	8.5\\
7.5	8.5\\
}--cycle;
\node[above, align=center, font=\bfseries\color{black!60!green}]
at (axis cs:8,8) {90.6\%};
\node[below, align=center, font=\bfseries\color{black!60!red}]
at (axis cs:8,8) {9.4\%};
\addplot [color=darkgray, line width=2.0pt, forget plot]
  table[row sep=crcr]{%
7.5	0.5\\
7.5	8.5\\
};
\addplot [color=darkgray, line width=2.0pt, forget plot]
  table[row sep=crcr]{%
0.5	7.5\\
8.5	7.5\\
};
\end{axis}
\end{tikzpicture}%

%% file: main.bbl
\begin{thebibliography}{10}
\providecommand{\url}[1]{#1}
\csname url@samestyle\endcsname
\providecommand{\newblock}{\relax}
\providecommand{\bibinfo}[2]{#2}
\providecommand{\BIBentrySTDinterwordspacing}{\spaceskip=0pt\relax}
\providecommand{\BIBentryALTinterwordstretchfactor}{4}
\providecommand{\BIBentryALTinterwordspacing}{\spaceskip=\fontdimen2\font plus
\BIBentryALTinterwordstretchfactor\fontdimen3\font minus
  \fontdimen4\font\relax}
\providecommand{\BIBforeignlanguage}[2]{{%
\expandafter\ifx\csname l@#1\endcsname\relax
\typeout{** WARNING: IEEEtran.bst: No hyphenation pattern has been}%
\typeout{** loaded for the language `#1'. Using the pattern for}%
\typeout{** the default language instead.}%
\else
\language=\csname l@#1\endcsname
\fi
#2}}
\providecommand{\BIBdecl}{\relax}
\BIBdecl

\bibitem{lyons1977semantics}
J.~Lyons, \emph{Semantics}.\hskip 1em plus 0.5em minus 0.4em\relax Cambridge
  University Press, 1977.

\bibitem{darvish2018flexible}
D.~Kourosh, W.~Francesco, B.~Barbara, S.~Enrico, M.~Fulvio, and C.~Giuseppe,
  ``Flexible human–robot cooperation models for assisted shop-floor tasks,''
  \emph{Mechatronics}, vol.~51, pp. 97--114, 2018.

\bibitem{yang2007gesture}
H.-D. Yang, A.-Y. Park, and S.-W. Lee, ``Gesture spotting and recognition for
  human--robot interaction,'' \emph{Transactions on Robotics}, vol.~23, no.~2,
  pp. 256--270, 2007.

\bibitem{iengo2014continuous}
S.~Iengo, S.~Rossi, M.~Staffa, and A.~Finzi, ``Continuous gesture recognition
  for flexible human-robot interaction,'' in \emph{Proceeding of the 2014 IEEE
  International Conference on Robotics and Automation (ICRA)}, Hong Kong,
  China, June 2014, pp. 4863--4868.

\bibitem{xie2016accelerometer}
R.~Xie and J.~Cao, ``Accelerometer-based hand gesture recognition by neural
  network and similarity matching,'' \emph{Sensors Journal}, vol.~16, no.~11,
  pp. 4537--4545, 2016.

\bibitem{khan2013exploratory}
A.~M. Khan, M.~H. Siddiqi, and S.-W. Lee, ``Exploratory data analysis of
  acceleration signals to select light-weight and accurate features for
  real-time activity recognition on smartphones,'' \emph{Sensors}, vol.~13,
  no.~10, pp. 13\,099--13\,122, 2013.

\bibitem{moazen2016airdraw}
D.~Moazen, S.~A. Sajjadi, and A.~Nahapetian, ``Airdraw: Leveraging smart watch
  motion sensors for mobile human computer interactions,'' in \emph{Proceedings
  of the 2016 IEEE Consumer Communications \& Networking Conference (CCNC)},
  Las Vegas, USA, January 2016, pp. 442--446.

\bibitem{bruno2014using}
B.~Bruno, F.~Mastrogiovanni, A.~Saffiotti, and A.~Sgorbissa, ``Using fuzzy
  logic to enhance classification of human motion primitives,'' in
  \emph{Proceeding of the 2014 International Conference on Information
  Processing and Management of Uncertainty in Knowledge-Based Systems (IPMU)},
  Montpellier, France, July 2014, pp. 596--605.

\bibitem{akl2010accelerometer}
A.~Akl and S.~Valaee, ``Accelerometer-based gesture recognition via
  dynamic-time warping, affinity propagation, \& compressive sensing,'' in
  \emph{Proceedings of the 2010 IEEE International Conference on Acoustics
  Speech and Signal Processing (ICASSP)}, Dallas, USA, March 2010, pp.
  2270--2273.

\bibitem{liu2009uwave}
J.~Liu, L.~Zhong, J.~Wickramasuriya, and V.~Vasudevan, ``uwave:
  Accelerometer-based personalized gesture recognition and its applications,''
  \emph{Pervasive and Mobile Computing}, vol.~5, no.~6, pp. 657--675, 2009.

\bibitem{wu2010hand}
X.-H. Wu, M.-C. Su, and P.-C. Wang, ``A hand-gesture-based control interface
  for a car-robot,'' in \emph{Proceedings of the 2010 IEEE International
  Conference on Intelligent Robots and Systems (IROS)}, Taipei, Taiwan, October
  2010, pp. 4644--4648.

\bibitem{khan2012gesthaar}
M.~Khan, S.~I. Ahamed, M.~Rahman, and J.-J. Yang, ``Gesthaar: An
  accelerometer-based gesture recognition method and its application in nui
  driven pervasive healthcare,'' in \emph{Proceedings of the 2012 IEEE
  International Conference on Emerging Signal Processing Applications (ESPA)},
  Las Vegas, USA, January 2012, pp. 163--166.

\bibitem{srivastava2016hand}
R.~Srivastava and P.~Sinha, ``Hand movements and gestures characterization
  using quaternion dynamic time warping technique,'' \emph{IEEE Sensors
  Journal}, vol.~16, no.~5, pp. 1333--1341, 2016.

\bibitem{porzi2013smart}
L.~Porzi, S.~Messelodi, C.~M. Modena, and E.~Ricci, ``A smart watch-based
  gesture recognition system for assisting people with visual impairments,'' in
  \emph{Proceedings of the 2013 ACM international workshop on Interactive
  multimedia on mobile \& portable devices (IMMPD)}, Bethesda, USA, June 2013,
  pp. 19--24.

\bibitem{shin2016dynamic}
S.~Shin and W.~Sung, ``Dynamic hand gesture recognition for wearable devices
  with low complexity recurrent neural networks,'' in \emph{Proceedings of the
  2016 IEEE International Symposium on Circuits and Systems (ISCAS)}, Montreal,
  Canada, May 2016, pp. 2274--2277.

\bibitem{bailador2007real}
G.~Bailador, D.~Roggen, G.~Tr{\"o}ster, and G.~Trivi{\~n}o, ``Real time gesture
  recognition using continuous time recurrent neural networks,'' in
  \emph{Proceedings of the 2007 ICST International Conference on Body area
  networks (BODYNETS)}, Florence, Italy, June 2007, p.~15.

\bibitem{coronado2017gesture}
E.~Coronado, J.~Villalobos, B.~Bruno, and F.~Mastrogiovanni, ``Gesture-based
  robot control: Design challenges and evaluation with humans,'' in
  \emph{Proceeding of the 2017 IEEE International Conference on Robotics and
  Automation (ICRA)}, Singapore, May 2017, pp. 2761--2767.

\end{thebibliography}
